\documentclass[conference]{IEEEtran}
\IEEEoverridecommandlockouts

\usepackage{cite}
\usepackage{booktabs}
\usepackage{amsmath,amssymb,amsfonts}
\usepackage{algorithmic}
\usepackage{graphicx}
\usepackage{textcomp}
\usepackage{xcolor}
\usepackage[colorlinks=true,linkcolor=blue,citecolor=blue]{hyperref}

\usepackage{url}
\usepackage{gensymb}
\usepackage{svg}
\usepackage{multirow}

\newcommand\ket[2][]{#1\lvert {#2} #1\rangle}

\usepackage[normalem]{ulem}
    
\begin{document}

\title{Expanding the Horizon: Enabling Hybrid Quantum Transfer Learning for Long-Tailed Chest X-Ray Classification}

\renewcommand{\thefootnote}{\roman{footnote}}

\author{\IEEEauthorblockN{Skylar Chan\IEEEauthorrefmark{1} \quad\quad Pranav Kulkarni\IEEEauthorrefmark{1} \quad\quad Paul H. Yi\IEEEauthorrefmark{2} \quad\quad Vishwa S. Parekh\IEEEauthorrefmark{1}}

\IEEEauthorblockA{\IEEEauthorrefmark{1}University of Maryland School of Medicine\\}

\IEEEauthorblockA{\IEEEauthorrefmark{2}St. Jude Children's Research Hospital\\}

\thanks{Corresponding author: vparekh@som.umaryland.edu}
\thanks{This work was supported by the University of Maryland School of Medicine, Baltimore, MD 21201.}
}

\maketitle

\begin{abstract}
Quantum machine learning (QML) has the potential for improving the multi-label classification of rare, albeit critical, diseases in large-scale chest x-ray (CXR) datasets due to theoretical quantum advantages over classical machine learning (CML) in sample efficiency and generalizability. While prior literature has explored QML with CXRs, it has focused on binary classification tasks with small datasets due to limited access to quantum hardware and computationally expensive simulations. To that end, we implemented a Jax-based framework that enables the simulation of medium-sized qubit architectures with significant improvements in wall-clock time over current software offerings. We evaluated the performance of our Jax-based framework in terms of efficiency and performance for hybrid quantum transfer learning for long-tailed classification across 8, 14, and 19 disease labels using large-scale CXR datasets. The Jax-based framework resulted in up to a 58\% and 95\% speed-up compared to PyTorch and TensorFlow implementations, respectively. However, compared to CML, QML demonstrated slower convergence and an average AUROC of 0.70, 0.73, and 0.74 for the classification of 8, 14, and 19 CXR disease labels. In comparison, the CML models had an average AUROC of 0.77, 0.78, and 0.80 respectively. In conclusion, our work presents an accessible implementation of hybrid quantum transfer learning for long-tailed CXR classification with a computationally efficient Jax-based framework. 
\end{abstract}


\section{Introduction}

Quantum machine learning (QML) has attracted recent interest due to rapid improvements in hardware and QML methods with theoretical quantum advantages over classical machine learning (CML) methods \cite{nmi2023seeking}. While quantum advantage typically refers to runtime improvements, potential advantages may also exist in sample efficiency, generalizability, expressibility, and trainability \cite{caro2022generalization,abbas2021power,huang2021power}. These advantages are particularly applicable for medical imaging tasks, where the growing scale of chest x-ray (CXR) datasets, coupled with the presence of rare, albeit critical, diseases labels results in deep learning (DL) models that are biased towards high-occurrence labels \cite{jia2023importance,holste2022long}.

While other groups have explored QML for training CXR DL models to detect diseases such as cardiomegaly, pneumonia, and COVID-19, their approaches were focused on binary classification tasks using small datasets that do not resemble the current state of CML with medical images \cite{decoodt2023hybrid,kulkarni2023classical,houssein2022hybrid}. More specifically, multiple diseases may be present in a patient simultaneously, thereby emphasizing the importance of detecting all potential diseases with multi-label classification. Furthermore, Bowles \emph{et al.} \cite{bowles2024better} suggest that it is not appropriate to extrapolate QML results from small datasets to large datasets.

The primary limitations of QML research beyond binary classification are limited access to quantum hardware, lack of integrated QML frameworks for medical imaging, and the large computational resources required for classical simulation \cite{asadi2024hybrid}. To that end, we developed a Jax-based software framework to lower the computational requirements for classical simulation of medium-sized qubit architectures on workstation-level hardware. We evaluated the Jax-based framework for long-tailed multi-label classification across 8, 14, and 19 disease labels using two large-scale CXR datasets. We further compared the performance of QML with CML to evaluate if the theoretical advantages of QML translate into empirical success for these tasks \cite{schuld2022quantum,arunachalam2018optimal,tang2022dequantizing,gil2024understanding,holmes2022connecting}.

Our contributions are two-fold:
\begin{enumerate}
    \item We develop an open-source Jax-based software framework for computationally efficient simulation of medium-sized qubit architectures on workstation-level hardware.
    \item We evaluate the scalability of hybrid quantum transfer learning for long-tailed multi-label classification tasks with large-scale CXR datasets.
\end{enumerate}

The relevant background for our study is provided in Section \ref{sec:background}. Our methods are detailed in Section \ref{sec:methods}. Implementation details of our Jax-based framework are provided in Section \ref{sec:implementation}. Our results are detailed in Section \ref{sec:results} and discussed in Section \ref{sec:discussion}.

\section{Background} \label{sec:background}

\subsection{Chest X-Ray Classification}

Chest X-Ray (CXR) is the most common medical imaging study for detecting life-threatening diseases like cardiomegaly and pneumonia \cite{ccalli2021deep}. Due to the high co-occurrence of findings within a single CXR, the ability to simultaneously detect the presence of multiple findings with multi-label classification is crucial. However, CXR classification is fundamentally different from natural image classification tasks because CXR datasets contain hundreds of thousands of high-resolution images visualizing anatomical structures, each potentially containing dozens of disease findings \cite{wang2017chestx,irvin2019chexpert,johnson2019mimic,bustos2020padchest}. These datasets are predicted to increase in size and resolution faster than improvements in hardware and software efficiency, as well as the number of registered radiologists in the USA \cite{jia2023importance}.

This rapid growth has led to an interesting problem where the high occurrence of CXRs that are healthy or contain common findings, coupled with CXRs containing rare yet critical findings, leads to a \emph{long-tailed distribution} of disease labels \cite{holste2022long}. Training on such datasets often results in DL models with performance that is biased towards high-occurrence labels. This disparity can lead to incorrect predictions for patients with rare long-tail diseases and disastrous downstream implications on patient health outcomes. The potential theoretical advantages of QML could translate to improved performance and efficiency in long-tailed CXR classifiers. Additionally, QML has primarily been evaluated on comparatively simpler classification problems such as binary or ternary classification on artificial data distributions and benchmark datasets like MNIST \cite{bowles2024better,sunkel2023hybrid,decoodt2023hybrid,kulkarni2023classical,houssein2022hybrid}. Thus, research on QML and long-tailed CXR classification can cooperatively benefit by using a more challenging problem to better understand where QML stands and how it can be potentially improved.

\subsection{Quantum Transfer Learning}

While there are methods to apply quantum computing to CML subroutines \cite{duan2020survey,liu2024towards} and quantum implementations of classical architectures \cite{havlivcek2019supervised,dallaire2018quantum,amin2018quantum}, the variational quantum circuit (VQC) has been extensively evaluated within and outside medical imaging due to its theoretical properties, simple implementation, and integration with backbone models through transfer learning \cite{cerezo2021variational}.

The VQC starts with $n$ qubits in some initial state, often prepared with data encoding methods, where classical data is encoded as quantum states. A series of parameterized quantum operations, such as rotation and entanglement gates, are then applied to these states. Finally, data is read out from quantum states into classical states. These circuits are trained by performing gradient descent to minimize a loss function of predictions obtained from postprocessing of the measurements of the quantum gradient, and are sometimes called ``quantum neural networks'' for this reason. Overall, the VQC embeds classical data into an intermediate quantum kernel, and projects quantum states back into classical data \cite{huang2021power}.

The choice of circuit design (ansatz) is critical for VQC performance. While VQCs tend to avoid exploding gradients \cite{farhi2018classification}, they can be prone to barren plateaus \cite{mcclean2018barren}. Furthermore, different ansatz also have a trainability and expressibility trade-off \cite{sim2019expressibility}, so the optimal ansatz for a classification problem must be carefully chosen. The design of the transferred model may also be important regarding the trainability of the model \cite{friedrich2022avoiding}.

The abstraction of classical input and classical output with VQCs enables hybrid classical-quantum models using transfer learning \cite{mari2020transfer}. The combination of a backbone model and classical preprocessing layers, followed by a VQC and classical postprocessing layers, is collectively referred to as a \emph{Dressed Quantum Circuit} (DQC). While transfer learning potentially obscures the impact of quantum layers, it reduces the computational resources required for training models with large-scale datasets, and the qubit requirements for the intermediate scale circuits necessary for multi-label classification.

\subsection{Computational Overhead of Quantum Simulation}

Prior work has indicated a need to scale up QML experiments to larger scales to determine whether QML behavior is consistent with smaller scales \cite{bowles2024better,chen2021end}. While the ultimate goal of quantum simulation is to demonstrate proof-of-concept scalability to real quantum hardware, VQCs are highly sensitive to quantum noise and errors \cite{stilck2021limitations,wang2021noise}. Work remains to implement fault-tolerant quantum error correction \cite{gottesman2022opportunities}, and quantum error mitigation may be limited by theoretical superpolynomial bounds on the number of samples needed to estimate expectation values commonly used in VQC optimization \cite{quek2022exponentially}. In the context of multi-label CXR classifiers, even small levels of noise could result in incorrect classifications and adversely impact patient health outcomes. This limits the benefits of efficient noisy simulation \cite{zhou2020limits,endo2021hybrid}. Thus, noiseless (ideal) quantum simulation is currently the most reliable and accessible approach to experimenting with QML.

However, the scalability of QML simulation is computationally intensive because the number of neurons in the output layer of DL models for multi-label classification is at least the number of disease labels to be classified. Given a fixed feature extraction backbone and a linear layer of size $n$ for classification, the scalability of computations is dependent on matrix multiplications between the size of the final layer of the feature extraction backbone and the size of the classification head. While matrix multiplication is slightly less than cubic in runtime complexity ($O(n^{\log_2 7})$ by Strassen's algorithm), since only a single dimension is variable (the length of the classification head), the scalability of a multi-label CML classifier is $O(n)$. For DQCs, the number of qubits required is at least the number of disease labels. Assuming a classical preprocessing linear layer of $n$ labels and an identically sized classical postprocessing linear layer, the scalability of the classical portion of the network is still $O(n)$ by dropping constant factors. However, a DQC with $n$ qubits requires a holding in memory a quantum state vector with $2^n$ length \cite{zhou2020limits}. Thus, the memory usage and compute time of a multi-label DQC classifier scales exponentially at a rate of $O(2^n)$\footnotemark.

\footnotetext{An alternative approach to quantum simulation is tensor networks, which are parallelizable and have linear memory scalability. However, finding the optimal contraction path for a tensor network is NP-hard, thus tensor network simulation still has fundamental scalability limits \cite{vallero2024state}.}

Although the wall-clock time of quantum simulation may deviate from this theoretical runtime analysis, the computational overhead of scalability in quantum systems have been major limitations in the scalability of prior work in QML inside and outside medical imaging. While experiments have managed to achieve 16-qubit simulations using multiple GPUs\cite{kyriienko2022unsupervised,gianelle2022quantum}, it should be possible to achieve this scale on a single GPU, thereby lowering the computational barrier of entry to QML research in medical imaging.

\section{Methods} \label{sec:methods}

\begin{figure}[!b]
    \centering
    \includegraphics[width=\linewidth]{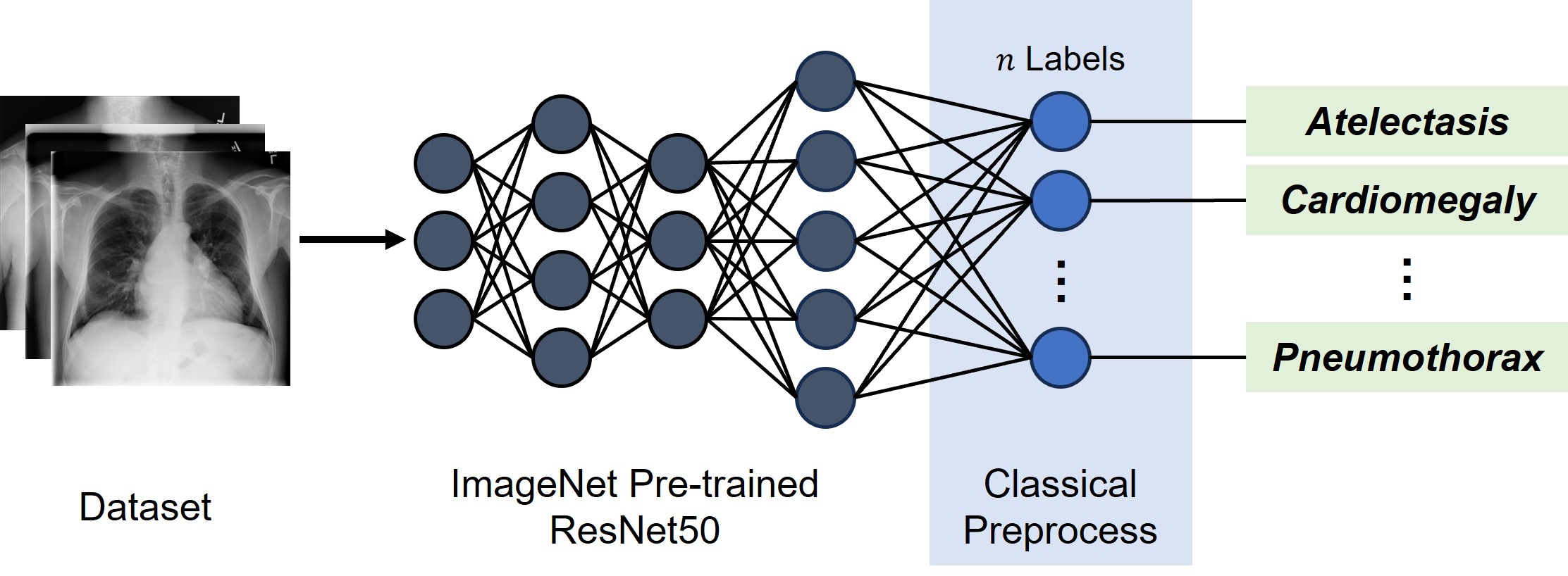}
    \caption{Classical deep learning model. Image features are extracted with ResNet50 and preprocessed with a linear layer before obtaining predictions.}
    \label{fig:cdl_overview}
\end{figure}

\begin{figure*}[!t]
    \centering
    \includegraphics[width=\linewidth]{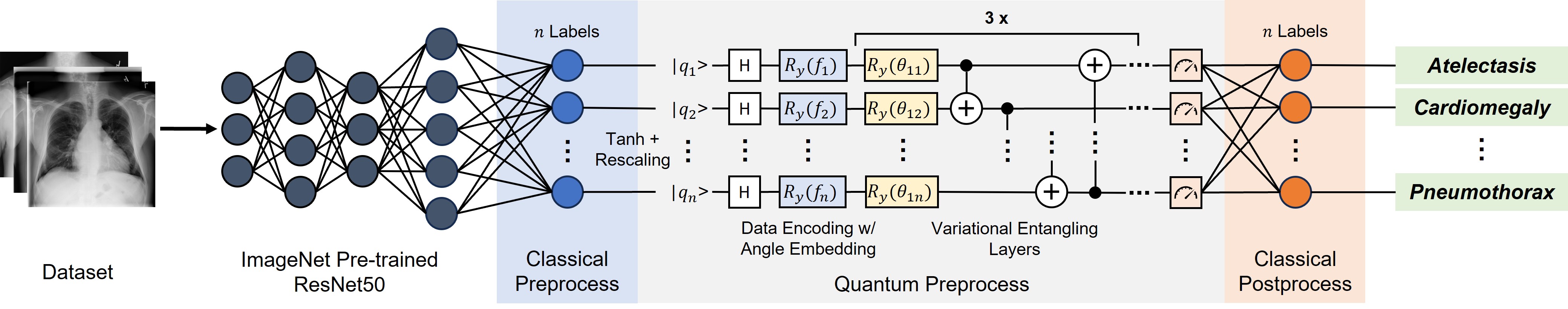}
    \caption{Dressed quantum circuit model. Image features are extracted with ResNet50, preprocessed down to size with a linear layer, then embedded into the quantum circuit with angle encoding applied to a 50/50 superposition of $\ket{0}$ and $\ket{1}$ after the Hadamard gate. Variational parameters (yellow) and CNOT gates (white) are applied, and measurements are fed into the classical postprocessing layer to obtain predictions. Preprocessing layers are highlighted in blue, variational quantum parameters in yellow, and postprocessing layers in orange.}
    \label{fig:dqc_overview}
\end{figure*}

\subsection{Model Architecture}

We implemented a DL model architecture consisting of a feature extractor and a classification head. For the feature extractor, we used an ImageNet pre-trained ResNet50 model from Hugging Face \cite{wolf2019huggingface}. For the classification head, we considered a classical and quantum approach. Parameter counts for each model are provided in Table \ref{tab:params}.

\subsubsection{Classical Deep Learning (CDL)}

After the feature extractor, we applied a classical preprocessing layer as the classification head (Fig. \ref{fig:cdl_overview}). It consisted of a linear fully-connected layer of size $n$, equal to the number of disease labels, followed by sigmoid activation.

\subsubsection{Dressed Quantum Circuit (DQC)}

After the feature extractor, we applied a Dressed Quantum Circuit (DQC) as the quantum classification head \cite{mari2020transfer,bowles2024better} (Fig. \ref{fig:dqc_overview}). The quantum circuit starts with $n$ qubits in the 0-state of the computational basis ($\ket{0}$). 
Hadamard gates are applied to each qubit to obtain a 50\% superposition of $\ket{0}$ and $\ket{1}$ as follows:
\begin{equation}
    H\ket{0} = \frac{1}{2}\ket{0} + \frac{1}{2}\ket{1} 
\end{equation}
The classification function of the DQC classifier $f: \mathbb{R}^m \to \mathbb{R}^n$ is defined as follows:
\begin{align}
    \hat{y} &= f(x; \theta, w_{in}, w_{out}) \\
    &= f_{out}(w_{out}, f_Q (\theta, f_{in} (w_{in}, x)))
\end{align}
where $f_{in}: \mathbb{R}^m \to \mathbb{R}^n$ is the classical preprocessing layers with weights $w_{in} \in \mathbb{R}^{m\times n}$, $f_Q: \mathbb{R}^n \to \mathbb{R}^n$ is the VQC with depth $d$ and trainable angles $\theta_{ij} \in \mathbb{R}^{n \times d}$, and $f_{out}: \mathbb{R}^n \to \mathbb{R}^n$ is the classical postprocessing layer with weights $w_{out} \in \mathbb{R}^{n \times n}$.

After the feature extractor, we apply a classical preprocessing layer (a linear layer of size $n$), followed by tanh activation and rescaling to $\left[-\frac{\pi}{2}, +\frac{\pi}{2}\right]$. Each output of the preprocessing layer is angle-embedded with a $R_Y$ rotation gate to its corresponding qubit. After the embedding, 3 layers of parameterized $R_Y$ gates followed by $CNOT$ gates are applied\footnotemark. Then, the noise-free expectation value of each qubit is measured. The measurements are passed to a postprocessing linear layer of size $n$. As the number of qubits is equal to the number of labels, we apply sigmoid activation to the postprocessing layer to obtain the final image classifications.

\footnotetext{Although 3 layers were chosen arbitrarily, prior work has indicated a potential lack of sensitivity to the number of layers for DQCs in binary classification \cite{azevedo2022quantum}.}

\subsection{Implementation} \label{sec:implementation}

Our open-source Jax-based implementation is available at \url{https://github.com/BioIntelligence-Lab/QuMI}. We used Python (version 3.12), CUDA (version 12), and other version-compatible libraries obtained from Conda-Forge. Further details on software dependencies and versions are available in the source code.

To shorten wall-clock training time, we used Jax \cite{bradbury2018jax}, which uses just-in-time (JIT) compilation to accelerate linear algebra operations for deep learning and quantum simulation on GPUs. Our implementation provides a QML pipeline for medical image analysis that uses Jax and Pennylane \cite{bergholm2018pennylane} to reduce the computational resources required to simulate medium-sized qubit systems on workstation-level hardware. 

We implemented the DL models using Flax \cite{heek2020flax} and the following libraries from the Jax ecosystem \cite{deepmind2020jax}: Optax for optimizers\footnote{https://github.com/google-deepmind/optax/}, Pix for image augmentations\footnote{https://github.com/google-deepmind/dm-pix/}, and Orbax for checkpointing\footnote{https://github.com/google/orbax/}. For quantum simulation, we used Pennylane to easily experiment with different DL libraries. Pennylane circuits were optimized using the \texttt{default.qubit.jax} device and the \texttt{best} differentiation method (backpropagation on our hardware).

\begin{figure}[!t]
    \centering
    \includegraphics[width=\linewidth]{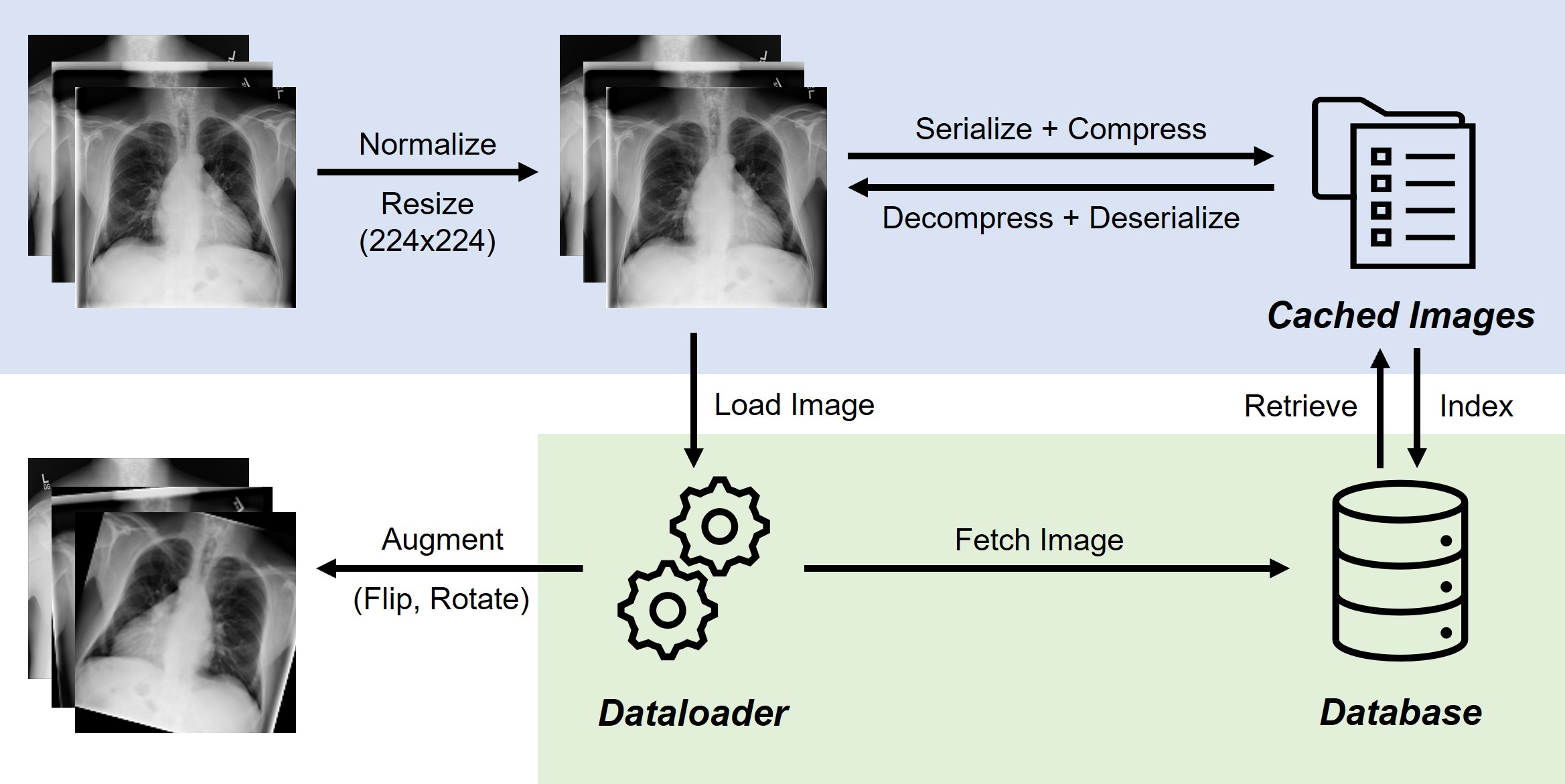}
    \caption{Dataloader implemented for experiments. Preprocessed images are serialized, compressed, and cached on disk ahead of time. At runtime, they are retrieved from the cache, decompressed, and deserialized. Image augmentations occur at runtime. A SQLite database tracks the file paths of the cached images.}
    \label{fig:preprocess}
\end{figure}

To avoid potential CUDA dependency conflicts, we implemented a framework-agnostic multithreaded caching dataloader that supports image preprocessing and augmentations (Fig. \ref{fig:preprocess}). Images were preprocessed with the Hugging Face \texttt{AutoImagePreprocessor} that was used to train the ResNet50 backbone. The images were serialized with SafeTensors \cite{patry_safetensors} (a non-executable binary format), compressed with Zstandard \cite{collet2018zstandard}, and cached with Diskcache \cite{jenks_diskcache}. Image augmentations were not cached and were performed dynamically after retrieving the cached preprocessed image. The dataloader was multithreaded using Joblib \cite{joblib_joblib}.

\section{Experiments} \label{sec:experiments}

\subsection{Benchmarks}

Pennylane computes gradients of quantum circuits by implementing autodifferention with Jax, PyTorch, and TensorFlow backends. To determine the scale of our implementation's performance gains, we benchmark wall-clock time (in s) of the training step of functionally equivalent  CDL and DQC models for each of these backends. All models were JIT compiled except for PyTorch due to lack of support for compiled quantum circuits. For each DL library, we load a single batch of zero-images and zero-labels (zero-vectors of the correct shape) into GPU memory to eliminate potential I/O and concurrency effects. We then perform 10 training steps to account for potential caching and compilation steps that may affect runtimes. Finally, we measure and compare the runtime of 30 additional training steps.

\subsection{Datasets}

\subsubsection{NIH-CXR-LT} 

The dataset consists of 19 disease labels with $n=112,120$ frontal CXRs from 30,805 patients \cite{wang2017chestx,holste2022long}. We use the provided training (70\%, $n=78,605$), validation (10\%, $n=12,535$), and testing (20\%, $n=21,081$) splits to train multi-label QML models.

\subsubsection{MIMIC-CXR-LT}

The dataset consists of the same 19 disease labels as the NIH-CXR-LT dataset with $n=377,110$ CXRs from 65,379 patients \cite{johnson2019mimic}. We use the provided testing split with lateral CXRs discarded ($n=48,860$) as our external test set to evaluate model generalizability.

\subsection{Classification Tasks}

To evaluate the scalability of QML for multi-label classification, we considered the following three sets of disease labels:
\begin{itemize}
    \item CXR-8: Atelectasis, Cardiomegaly, Effusion, Infiltration, Mass, Nodule, Pneumonia, and Pneumothorax.
    \item CXR-14: All labels from CXR-8, with the addition of Consolidation, Edema, Emphysema, Fibrosis, Pleural Thickening, and Hernia.
    \item CXR-19: All labels from CXR-14, with the addition of Calcification of the Aorta, Pneumomediastinum, Pneumoperitoneum, Subcutaneous Emphysema, and Tortuous Aorta.
\end{itemize}

\begin{table}[!t]
    \centering
    \caption{Number of parameters by component.}
    \label{tab:params}
    \begin{tabular}{lcc} \toprule
         \textbf{Component} & \textbf{\# Labels} & \textbf{\# Parameters}\\ \midrule
         ResNet50 & - & 23,508,032\\
         \multirow{3}{*}{Classical Preprocess}&  8&  16,392\\
         &  14&  28,686\\
         &  19&  38,931\\ \midrule
         \multirow{3}{*}{Quantum Preprocess}&  8&  24\\
         &  14&  42\\
         &  19&  57\\
         \multirow{3}{*}{Classical Postprocess}&  8&  72\\
         &  14&  210\\
         & 19&380\\ \bottomrule
    \end{tabular}
\end{table}

\begin{table}[!t]
    \centering
    \caption{Long-tailed distribution of disease labels from NIH-CXR-LT and MIMIC-CXR-LT by data split and sorted by occurrence.} 
    \label{tab:label_frequency_table}
    \begin{tabular*}{\linewidth}{@{\extracolsep{\fill}} llllll} \toprule
 \multirow{2}{*}{\textbf{Label}} & \multicolumn{4}{c}{\textbf{NIH}} & \multirow{2}{*}{\textbf{MIMIC}}\\ \cmidrule{2-5}
         &  \textbf{Train}&  \textbf{Val} &  \textbf{Test}& \textbf{Total}&  \\ \midrule
         Infiltration&  12739&  1996& 5159 & 19894& 1781\\
         Effusion&  7919&  1663& 3735 & 14586& 14586\\
         Atelectasis&  7587&  1272& 2700& 11559& 14014\\
         Nodule& 4359& 667 & 1305& 6331& 1309\\
         Mass& 3689& 764& 1329& 5782& 1006\\
         Pneumothorax& 2432& 764& 2106& 5302& 3282\\
         Consolidation&  2626&  544& 1497& 4667& 3232\\
         Pleural Thickening& 1998& 485& 902& 3385& 590\\
         Cardiomegaly&  1590&  318& 868& 2776& 14703\\
         Emphysema&  1327&  272& 917& 2516& 649\\
         Edema&  1283&  269& 751& 2303& 8444\\
         Subcutaneous Emphysema& 957& 221& 813& 1991& 572\\
         Fibrosis&  1138&  183& 365& 1686&  271\\
         Pneumonia& 806& 173& 452& 1431& 8242\\
         Tortuous Aorta& 598& 49& 95& 742& 547\\
         Calcification of the Aorta& 368& 32& 55& 455& 704\\
         Pneumoperitoneum& 214& 33& 69& 316& 136\\
         Pneumomediastinum& 88& 22& 143& 253& 176\\
         Hernia&  130&  35 & 62& 227& 739\\ \bottomrule
    \end{tabular*}
\end{table}

The classification tasks were chosen based on the different versions of the NIH dataset. The long-tailed distribution of both datasets is provided in Table \ref{tab:label_frequency_table}.

\subsection{Training Procedure}

We trained CDL and DQC models on CXR-8, CXR-14, and CXR-19 classification tasks using 5 random seeds. Backbone weights were initialized from the pre-trained ResNet50 checkpoint. Classical linear layer weights were randomly initialized using the default method (Lecun normal \cite{klambauer2017self}). Parameterized $R_Y$ gates of the DQC were initialized using a random normal distribution with a standard deviation of $2\pi$. Images were preprocessed ahead of time by resizing to 256 pixels on the shortest side, center-cropping to 224x224, and normalizing pixel values to ImageNet statistics. Training batches were shuffled and augmented during training with random horizontal flip ($p=0.5$) and random rotation ($\theta=\pm 15\degree$) \cite{holste2023does}. To ensure that CDL and DQC models with the same random seed use the same order of training examples, trials with the same random seed used the same pseudorandom permutations and augmentations for the training batches. Models were trained with a batch size of 32 to avoid running out of GPU memory, and optimized to minimize mean binary cross entropy (BCE) loss over all diseases labels using Adam (learning rate = 1e-4). Early stopping was used if validation loss did not improve after 5 epochs, with a maximum of 50 training epochs. The best performing model was determined by the model checkpoint with the lowest validation loss.

Models were trained in parallel over 4 NVIDIA RTX A6000 GPUs and Intel Xeon CPUs. All models used 4 CPU threads, except for DQCs, which used 1 or 2 threads to avoid running out of GPU memory.

\subsection{Metrics and Statistical Analysis}

To evaluate model trainability, we compared the convergence of the CDL and DQC models using the training and validation BCE loss across each step. To evaluate the performance of the CDL and DQC models, we measured the mean and per-label area under the receiver operating characteristic curve (AUROC) on the internal NIH and external MIMIC test sets across the CXR-8, CXR-14, and CXR-19 classification tasks. We compared the mean and per-label AUROCs of the CDL and DQC models using paired t-tests. Statistical significance was defined as $p<0.05$. 

To identify per-label performance trends across experiments, we generate volcano plots of these $p$-values against the percent difference in AUROC of CDL and DQC models, which is defined as follows:

\begin{equation}
    \text{\% diff. AUROC} = 2 * \frac{\text{AUROC}_{CDL} - \text{AUROC}_{DQC}}{\text{AUROC}_{CDL} + \text{AUROC}_{DQC}}
\end{equation}

More specifically, we produce scatter plots where each point represents an experiment between CDL and DQC on the same classification task, and the $-\log_{10}(p\text{-value})$ of the paired t-test is plotted against the $\log_{2}(\text{\% diff. AUROC})$. We include the labels in CXR-8 and the additional labels added in CXR-14, as these are present across multiple experiments.
\section{Results} \label{sec:results}

\subsection{Benchmarks}

Our benchmarks indicate that our Jax-based implementation is the fastest for simulating DQC across all three classification tasks when compared to functionally equivalent TensorFlow and PyTorch implementations (Table \ref{tab:benchmarks}). Extrapolated to an entire epoch of 2,454 batches:
\begin{itemize}
    \item For CXR-8, our implementation takes 2.45 mins/epoch and is 25\% faster than TensorFlow (3.27 mins/epoch) and 54\% faster than PyTorch (5.32 mins/epoch). 

    \item For CXR-14, our implementation takes 3.27 mins/epoch and is 81\% faster than TensorFlow (17.59 mins/epoch) and 56\% faster than PyTorch (7.36 mins/epoch). 

    \item For CXR-19, our implementation takes 32.72 mins/epoch and is 95\% faster than TensorFlow (612.27 mins/epoch) and 58\% faster than PyTorch (77.30 mins/epoch). 
\end{itemize}

\begin{table}[t!]
    \centering
    \caption{Benchmarked wall-clock runtimes (in s) between TensorFlow, PyTorch, and our Jax-based implementation across CXR-8, and CXR-14, and CXR-19 classification tasks for a zero-batch.}
    \label{tab:benchmarks}
    \begin{tabular*}{\linewidth}{@{\extracolsep{\fill}} lccc} \toprule
         & \textbf{TensorFlow} & \textbf{PyTorch} & \textbf{Jax}$^*$ \\ \midrule
        \multicolumn{4}{c}{\textbf{CXR-8}} \\ \midrule
        CDL & $0.06 \pm 0.00$ & $0.07 \pm 0.00$ & $0.06 \pm 0.00$ \\
        DQC & $0.08 \pm 0.00$ & $0.13 \pm 0.00$ & $0.06 \pm 0.00$ \\ \midrule
        \multicolumn{4}{c}{\textbf{CXR-14}} \\ \midrule
        CDL & $0.06 \pm 0.00$ & $0.07 \pm 0.00$ & $0.06 \pm 0.00$ \\
        DQC & $0.43 \pm 0.00$ & $0.18 \pm 0.00$ & $0.08 \pm 0.00$ \\ \midrule
        \multicolumn{4}{c}{\textbf{CXR-19}} \\ \midrule
        CDL & $0.06 \pm 0.00$ & $0.07 \pm 0.00$ & $0.06 \pm 0.00$ \\
        DQC & $14.97 \pm 0.01$ & $1.89 \pm 0.00$ & $0.80 \pm 0.00$ \\ \bottomrule
        \multicolumn{4}{l}{$^*$ Our implementation.} \\
    \end{tabular*}
\end{table}

\subsection{Model Trainability}

\begin{figure*}[!t]
    \centering
    \includegraphics[width=\linewidth]{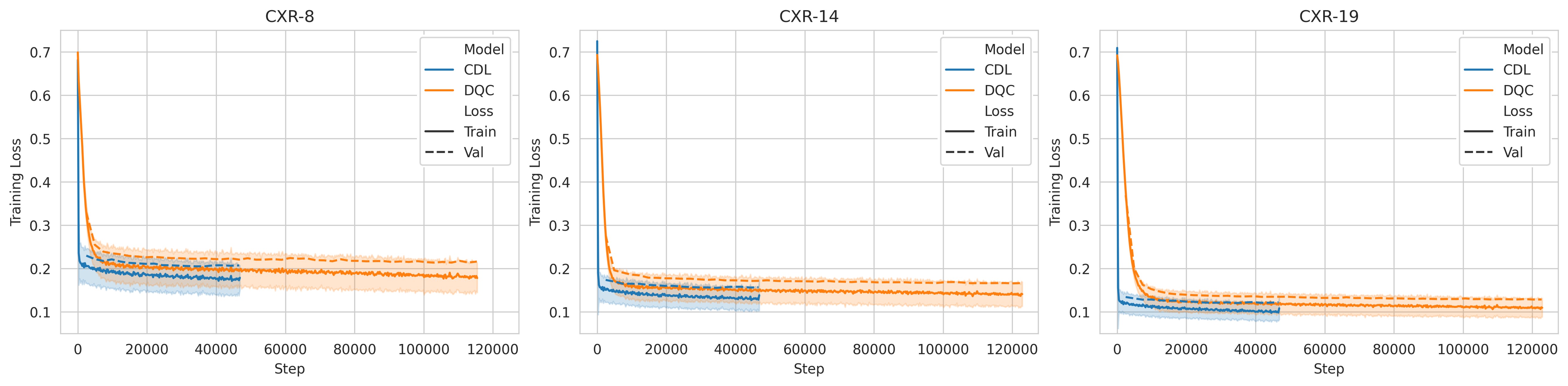}
    \caption{Training and validation loss between CDL and DQC models across CXR-8, CXR-14, and CXR-19 multi-label classification tasks.}
    \label{fig:trainability}
\end{figure*}

We observed that across the CXR-8, CXR-14, and CXR-19 classification tasks, the DQC models had a slower convergence rate compared to the CDL models, as shown in Figure \ref{fig:trainability}. This resulted in DQC models taking more than twice as many epochs to converge with early stopping. Moreover, the DQC models demonstrated a higher validation loss compared to CDL models across CXR-8 (0.22 vs 0.21), CXR-14 (0.17 vs 0.15), and CXR-19 (0.13 vs 0.12) classification tasks.

\subsection{Model Performance}

\begin{figure}[!t]
    \centering
    \includegraphics[width=0.8\linewidth]{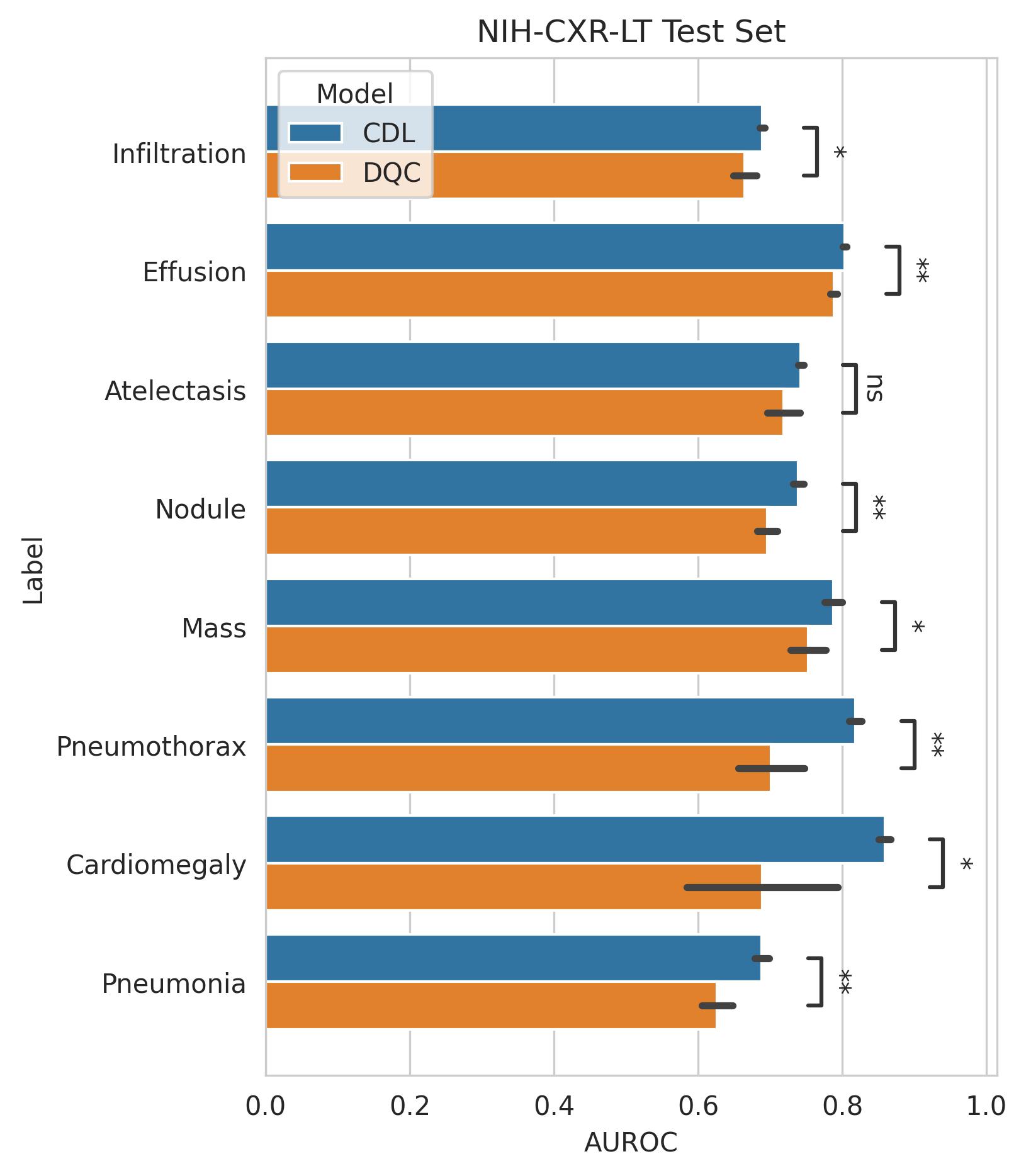}
    \caption{Mean per-label AUROC between CDL and DQC models on NIH-CXR-LT test set for CXR-8 classification task, sorted by occurrence.\\(ns: $p \ge 0.05$, *: $p<0.05$, **: $p<0.01$, ***: $p<0.001$).}
    \label{fig:main_results_nih_8}
\end{figure}

\begin{figure}[!t]
    \centering
    \includegraphics[width=0.9\linewidth]{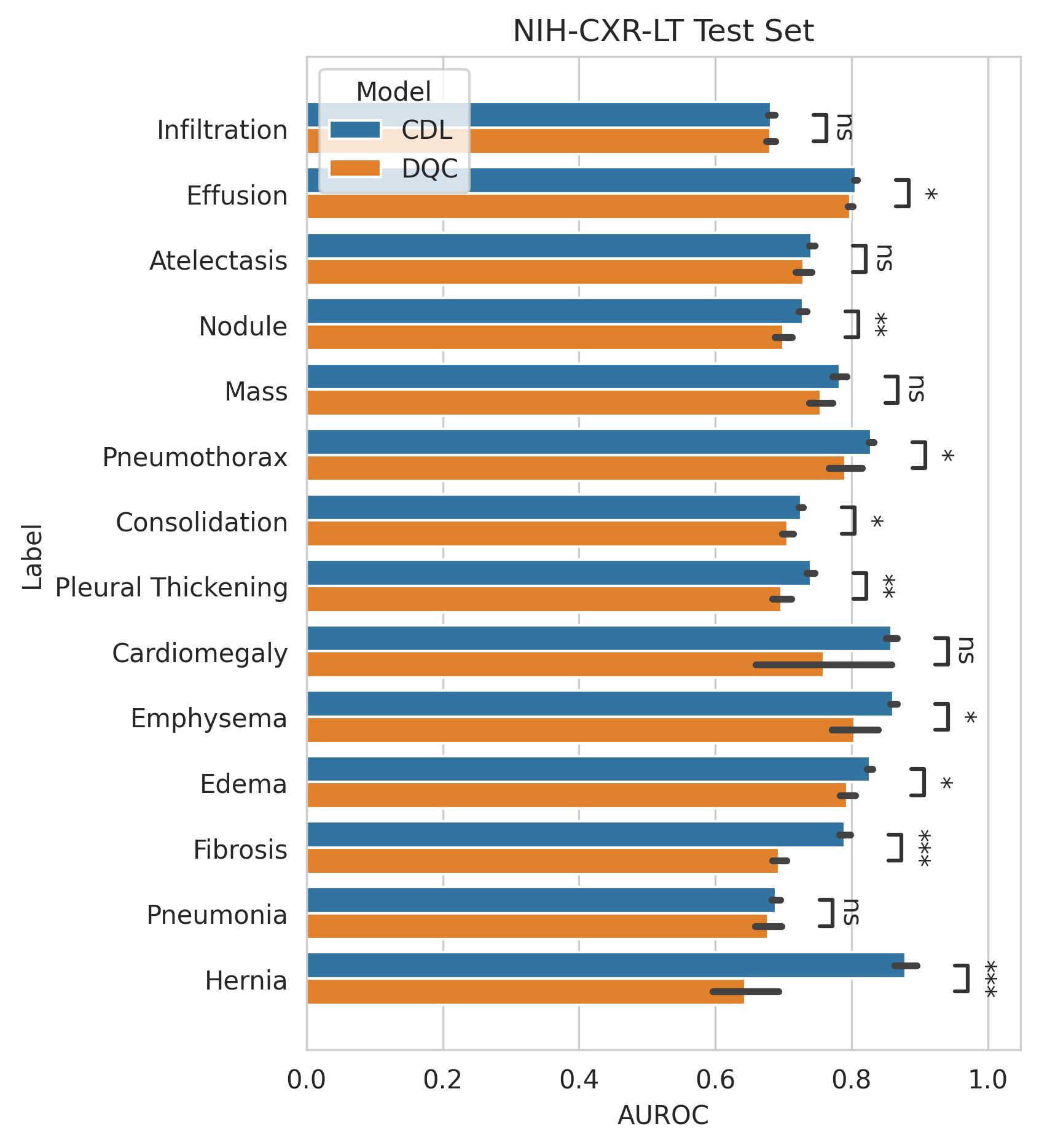}
    \caption{Mean per-label AUROC between CDL and DQC models on NIH-CXR-LT test set for CXR-14 classification task, sorted by occurrence.\\(ns: $p \ge 0.05$, *: $p<0.05$, **: $p<0.01$, ***: $p<0.001$).}
    \label{fig:main_results_nih_14}
\end{figure}

\begin{figure}[!t]
    \centering
    \includegraphics[width=\linewidth]{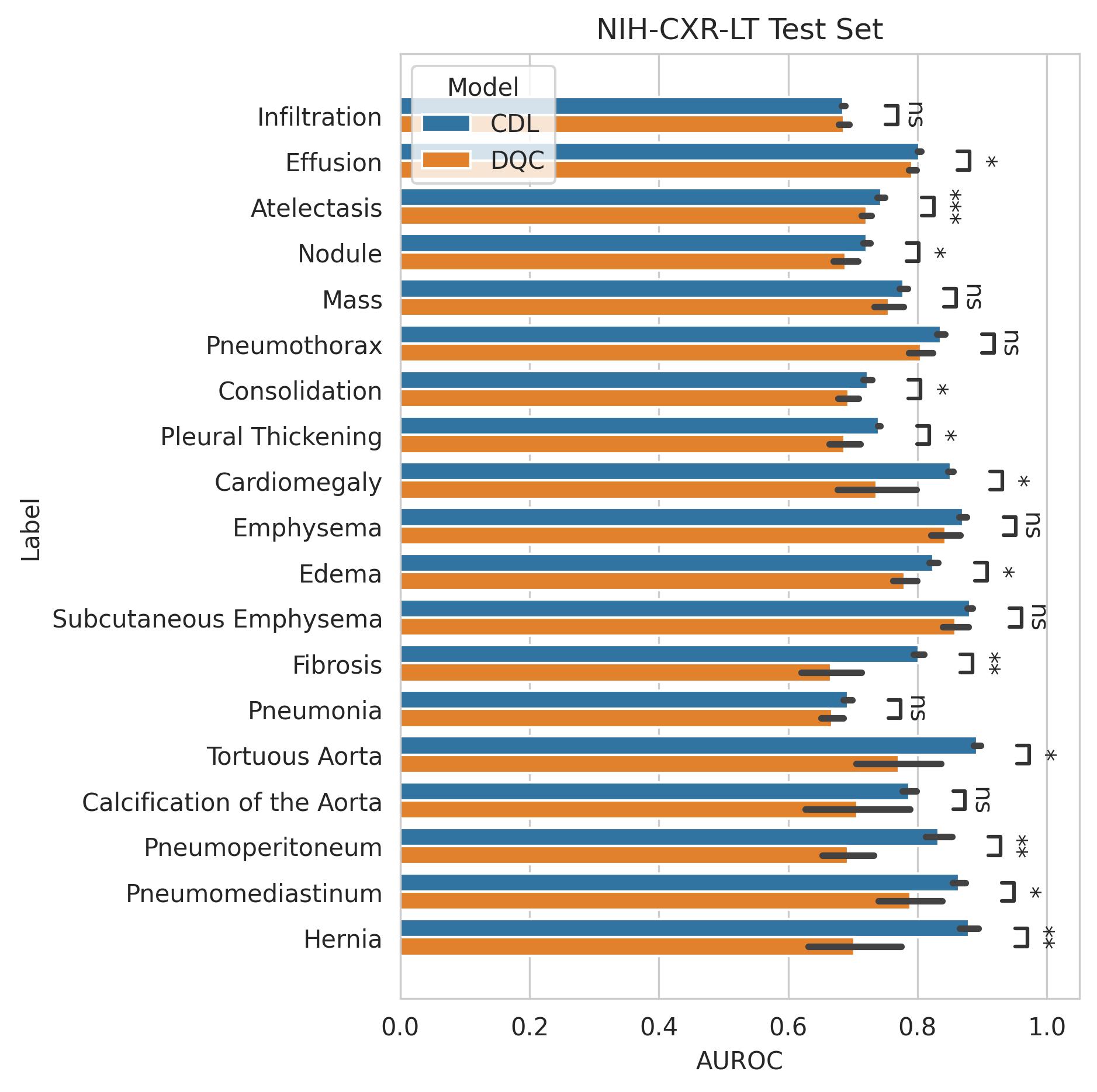}
    \caption{Mean per-label AUROC between CDL and DQC models on NIH-CXR-LT test set for CXR-19 classification task, sorted by occurrence.\\(ns: $p \ge 0.05$, *: $p<0.05$, **: $p<0.01$, ***: $p<0.001$).}
    \label{fig:main_results_nih_19}
\end{figure}

When comparing the model performance (mean AUROC) on the internal NIH test set, we observed that the CDL models significantly outperformed the DQC models by 9.5\%  for CXR-8 ($0.70 \pm 0.02$ vs $0.77 \pm 0.00$, $p=0.005$), by 6.6\% for CXR-14 ($0.73 \pm 0.02$ vs $0.78 \pm 0.00$, $p=0.004$), and by 7.8\% for CXR-19 ($0.74 \pm 0.03$ vs $0.80 \pm 0.00$, $p=0.009$). Figures \ref{fig:main_results_nih_8}, \ref{fig:main_results_nih_14}, and \ref{fig:main_results_nih_19} illustrate the label-wise comparison in performance between the CDL and DQC models. 

\begin{figure}[!t]
    \centering
    \includegraphics[width=0.8\linewidth]{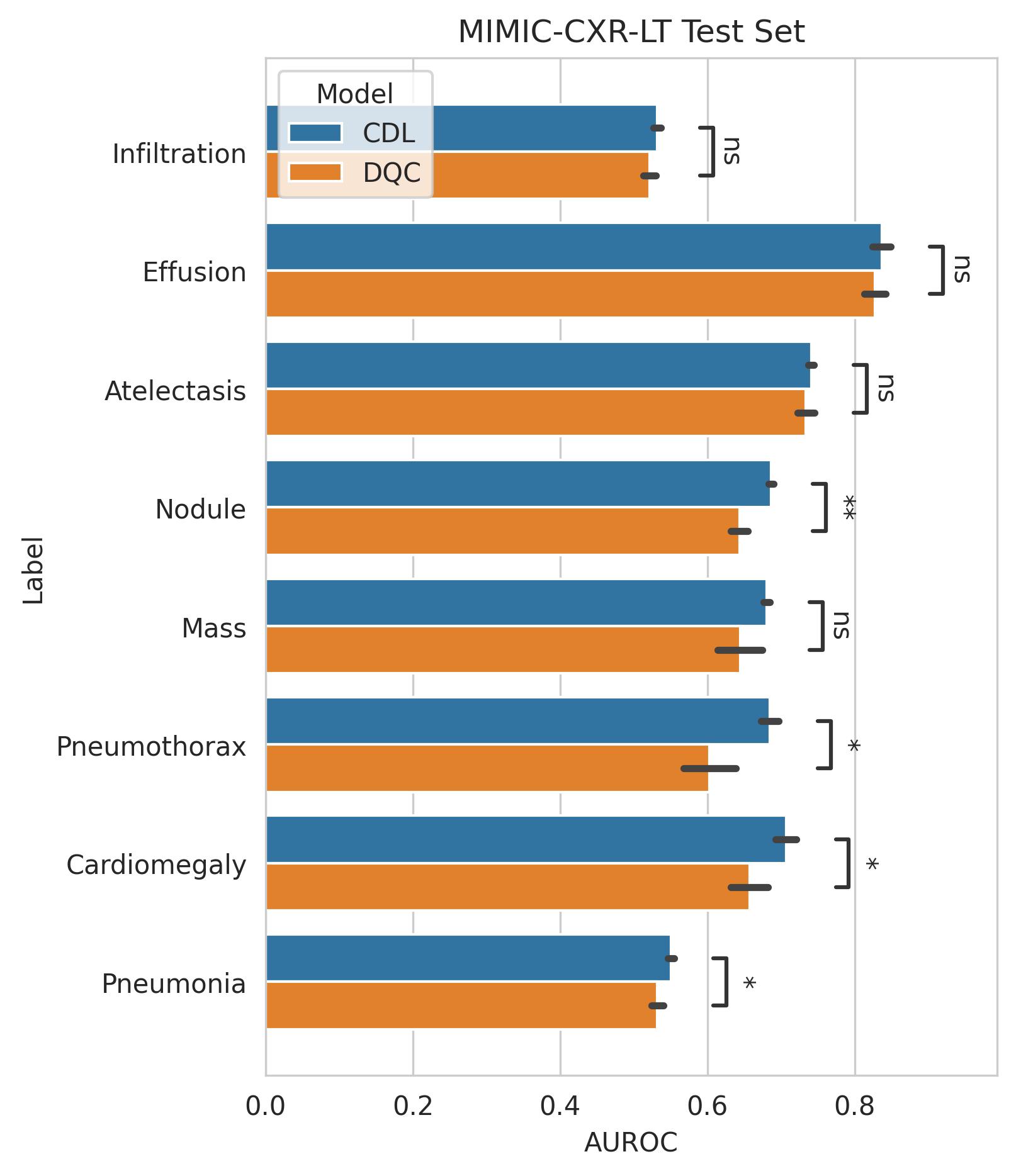}
    \caption{Mean per-label AUROC between CDL and DQC models on MIMIC-CXR-LT test set for CXR-8 classification task, sorted by occurrence.\\(ns: $p \ge 0.05$, *: $p<0.05$, **: $p<0.01$, ***: $p<0.001$).}
    \label{fig:main_results_mimic_8}
\end{figure}

\begin{figure}[!t]
    \centering
    \includegraphics[width=0.9\linewidth]{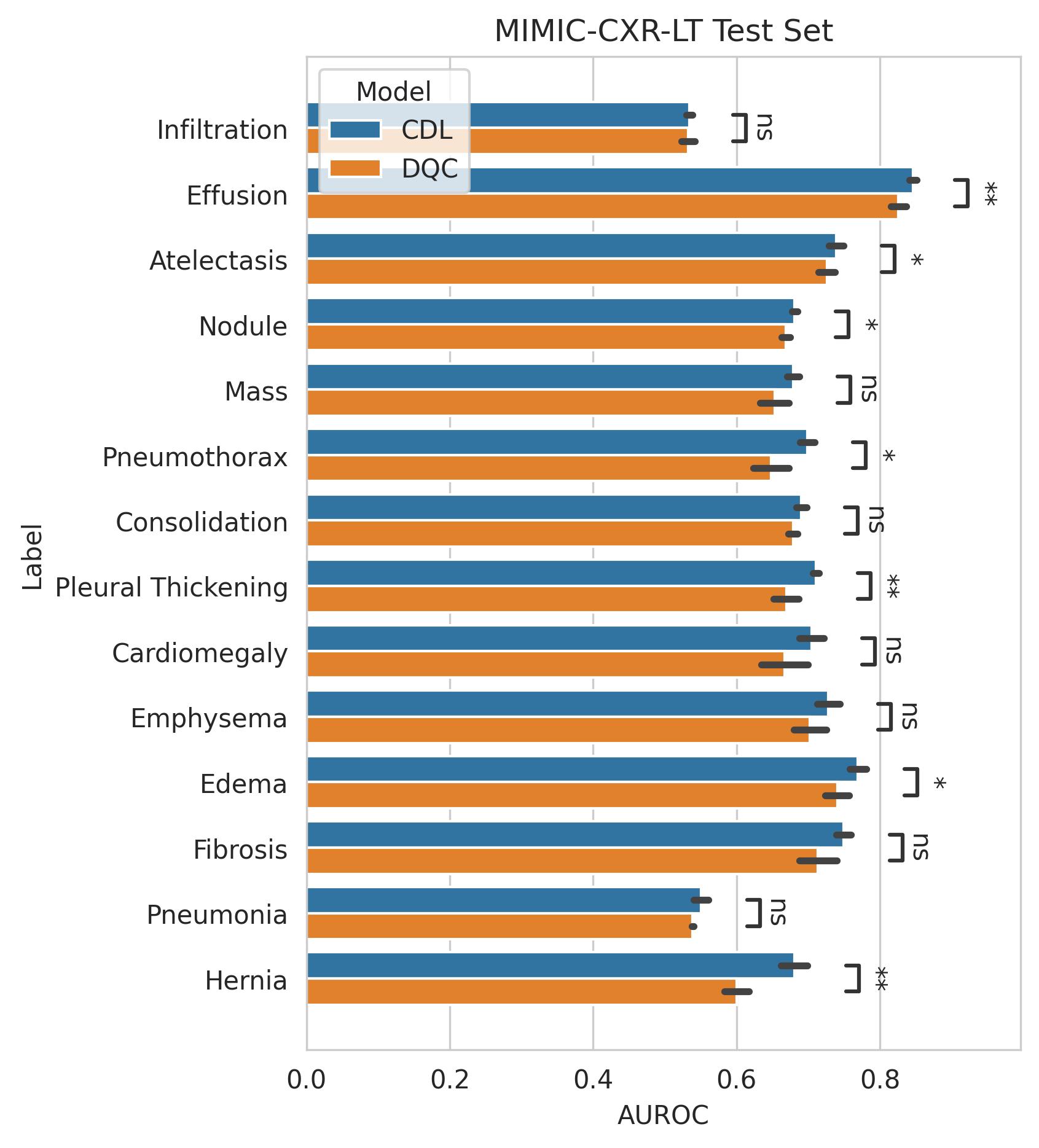}
    \caption{Mean per-label AUROC between CDL and DQC models on MIMIC-CXR-LT test set for CXR-14 classification task, sorted by occurrence.\\(ns: $p \ge 0.05$, *: $p<0.05$, **: $p<0.01$, ***: $p<0.001$).}
    \label{fig:main_results_mimic_14}
\end{figure}

\begin{figure}[!t]
    \centering
    \includegraphics[width=\linewidth]{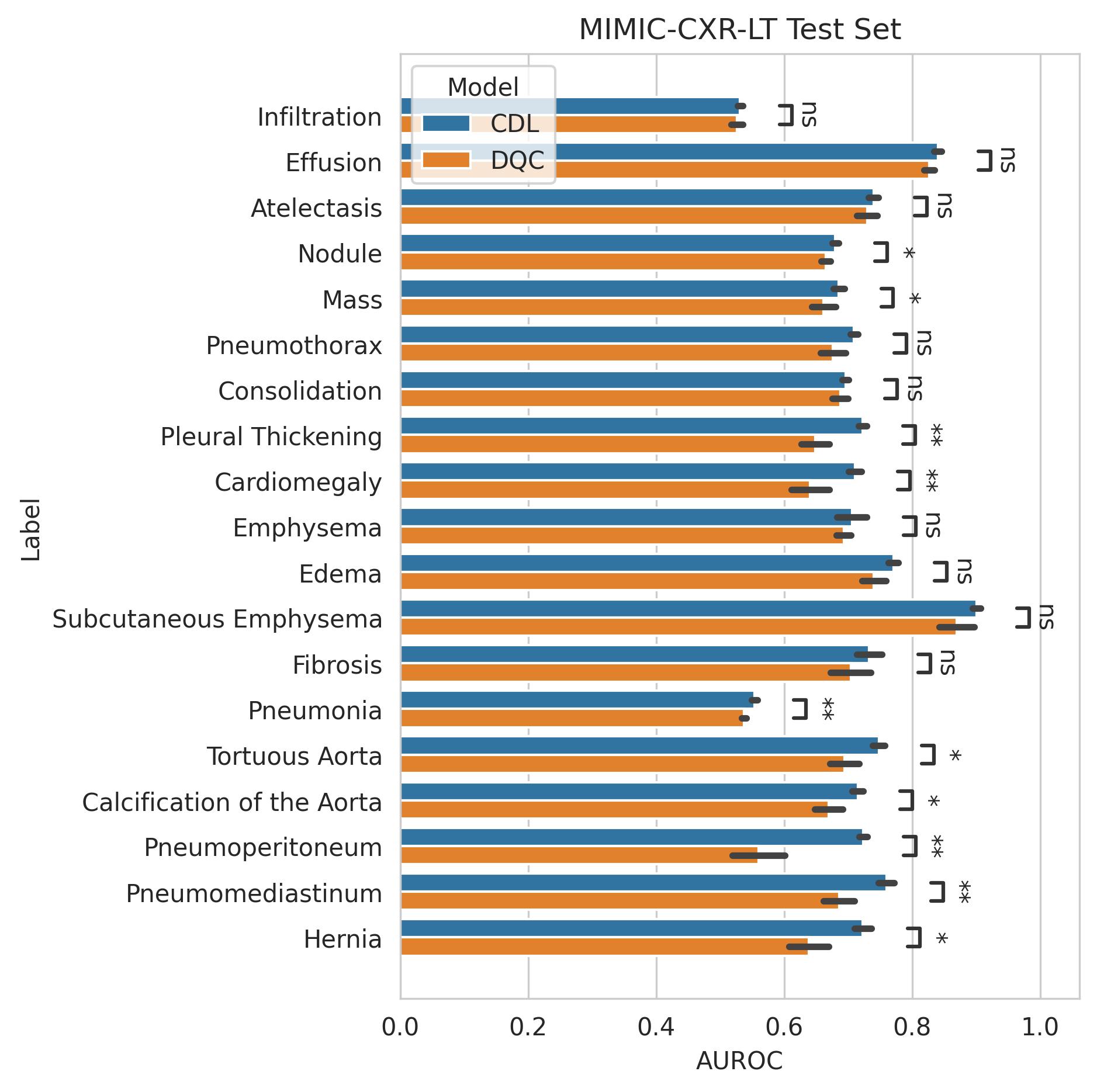}
    \caption{Mean per-label AUROC between CDL and DQC models on MIMIC-CXR-LT test set for CXR-19 classification task, sorted by occurrence.\\(ns: $p \ge 0.05$, *: $p<0.05$, **: $p<0.01$, ***: $p<0.001$).}
    \label{fig:main_results_mimic_19}
\end{figure}

When evaluating model generalizability on the external MIMIC test set, we observed that DQC models had a significantly lower mean AUROCs when compared to CDL models. However, the difference in performance between the two models was lower than the differences in the internal NIH test set, with a difference of 4.5\% for CXR-8 ($0.65 \pm 0.01$ vs $0.68 \pm 0.00$, $p=0.009$), 4.4\% for CXR-14 ($0.67 \pm 0.01$ vs $0.70 \pm 0.00$, $p=0.008$), and 5.7\% for CXR-19 ($0.68 \pm 0.01$ vs $0.72 \pm 0.00$, $p=0.002$). Figures \ref{fig:main_results_mimic_8}, \ref{fig:main_results_mimic_14}, and \ref{fig:main_results_mimic_19} illustrate this comparative performance between the two models, stratified by disease labels.

\begin{figure}[!t]
    \centering
    \includegraphics[width=\linewidth]{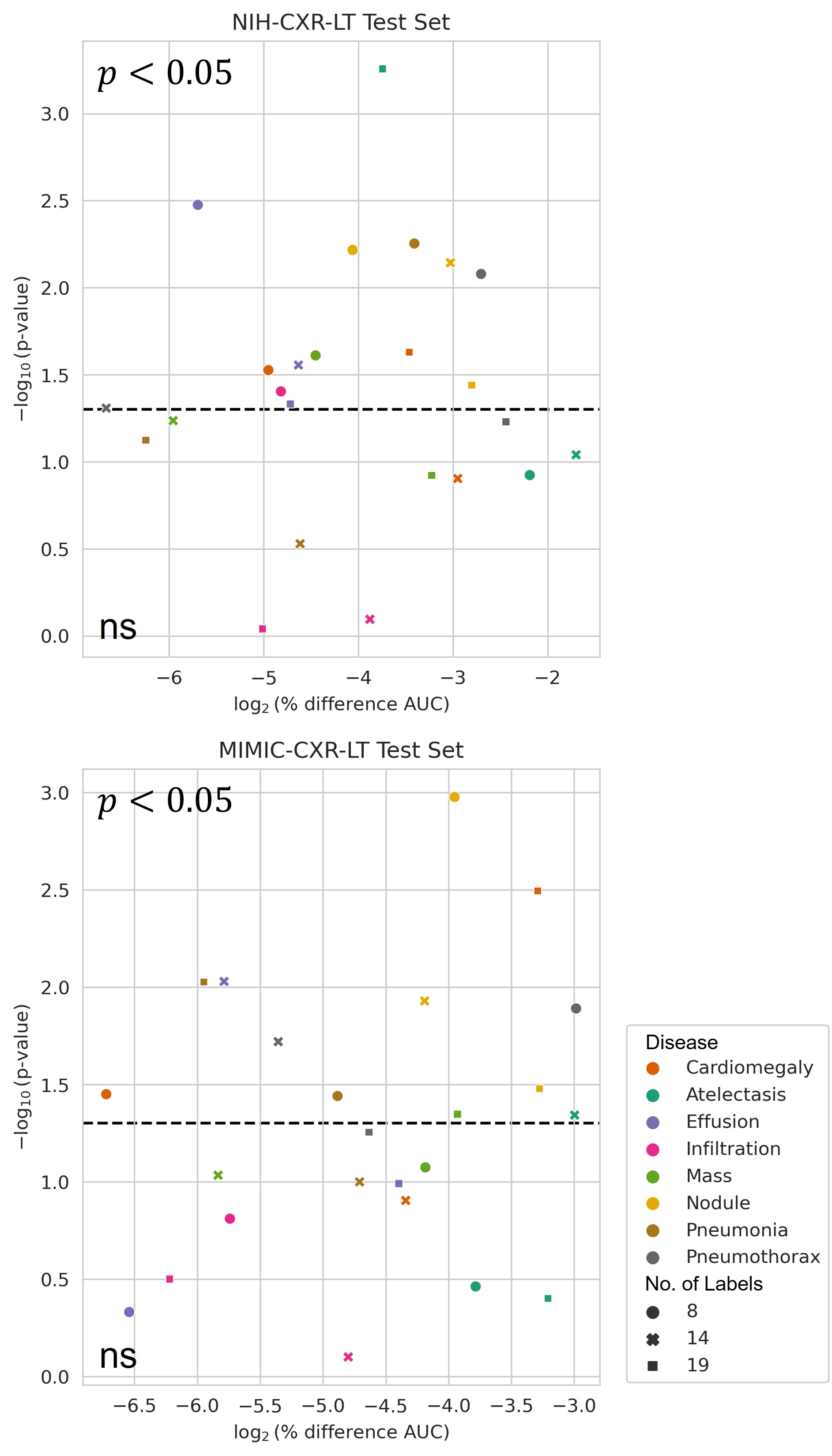}
    \caption{Volcano plot of performance on CXR-8 labels across experiments. Negative $\log_{10}$-transformed $p$-values of the paired t-tests for mean AUROC of CDL and DQC are plotted against the $\log_{2}$-transformed \% difference in AUROC between CDL and DQC. The dashed line marks the significance threshold ($p < 0.05$).}
    \label{fig:volcano_results_8}
\end{figure}

\begin{figure}[!t]
    \centering
    \includegraphics[width=\linewidth]{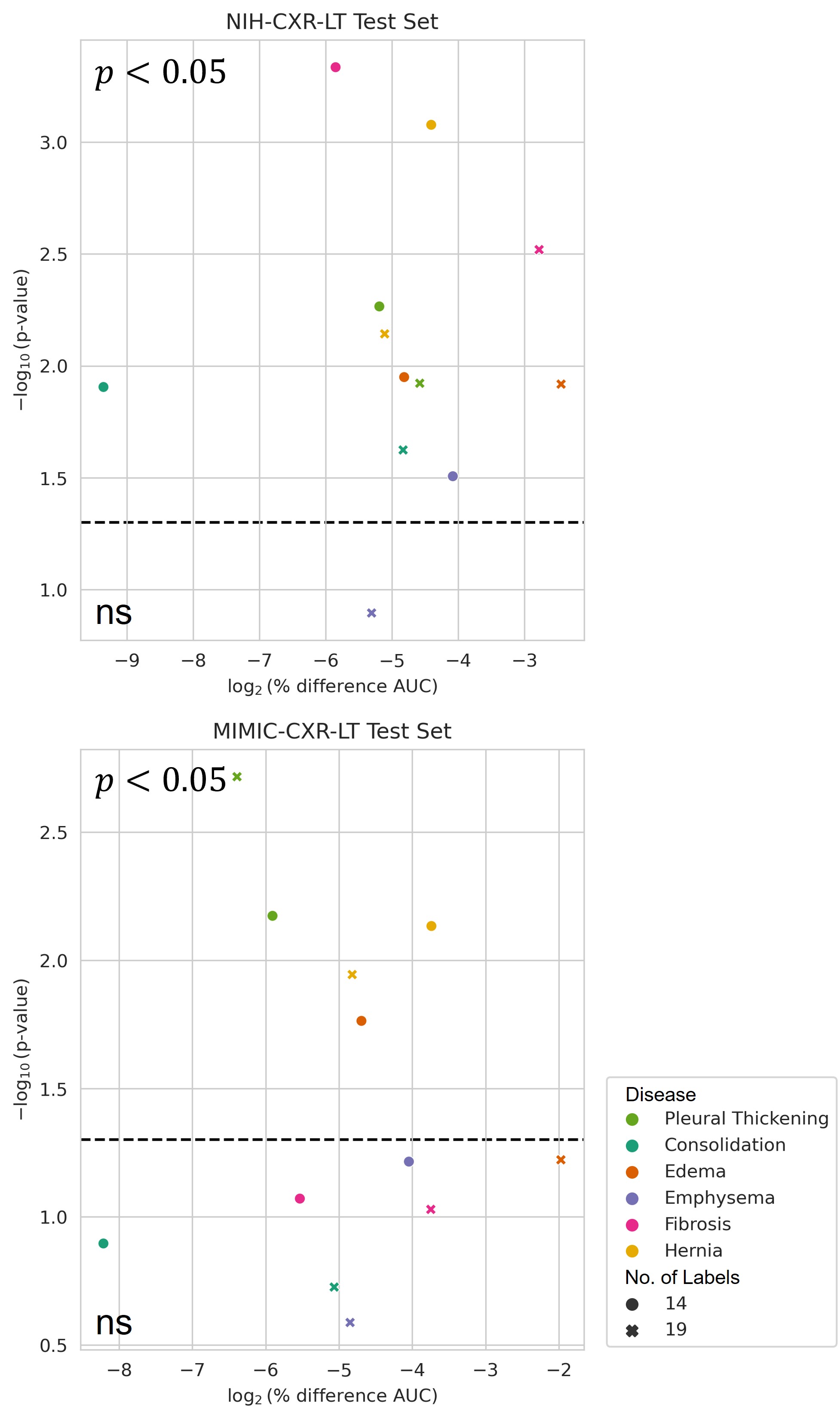}
    \caption{Volcano plot of performance on the CXR-14 additional labels across experiments. Negative $\log_{10}$-transformed $p$-values of the paired t-tests for mean AUROC of CDL and DQC are plotted against the $\log_{2}$-transformed \% difference in AUROC between CDL and DQC. The dashed line marks the significance threshold ($p < 0.05$).}
    \label{fig:volcano_results_14}
\end{figure}

Our volcano plots (Figures \ref{fig:volcano_results_8} and \ref{fig:volcano_results_14}) summarize per-label performance trends across paired models for CXR-8 and CXR-14 labels, respectively. Points above the dotted line are considered significant ($p < 0.05$), and points farther to the right have a higher (more positive) percent difference in AUC, thus indicating a stronger advantage of CDLs over DQCs. We observe that the 4 labels with the largest occurrence in the NIH dataset show the most consistent trends:

\begin{itemize}
    \item Infiltration had no significant \% diff. AUROC except for the CXR-8 task on the NIH test set.
    \item Effusion had significantly higher \% diff. AUROC except for the CXR-8 and CXR-19 tasks on the MIMIC test set.
    \item Atelectasis tended to exhibit the highest \% diff. AUROC among all labels. Although this difference was not statistically significant, it could potentially be considered clinically significant.
    \item Nodule had significantly higher \% diff. AUROC in DQCs compared to CDLs on both test sets.
\end{itemize}

For CXR-14 labels on the NIH test set, every additional label exhibited significantly higher \% diff. AUROC, except for Emphysema in the CXR-19 task. Conversely, for CXR-14 labels on the MIMIC test set, only Hernia exhibited significantly higher \% diff. AUROC for both CXR-14 and CXR-19 tasks.

\begin{figure}[!t]
    \centering
    \includegraphics[width=0.75\linewidth]{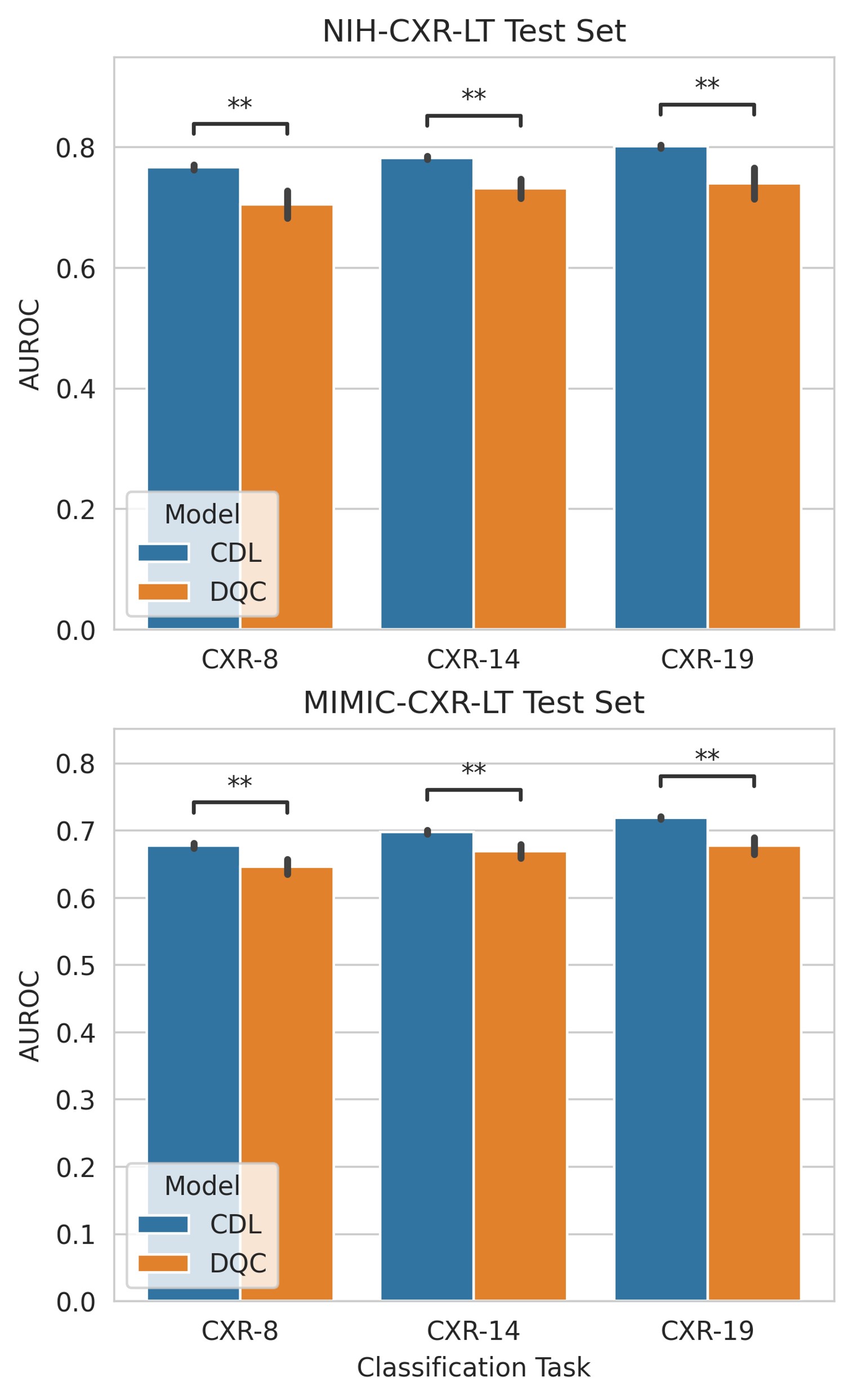}
    \caption{Mean AUROC between CDL and DQC models on NIH-CXR-LT and MIMIC-CXR-LT test sets across CXR-8, CXR-14, and CXR-19 classification tasks. (ns: $p>0.05$, *: $p<0.05$, **: $p<0.01$, ***: $p<0.001$).}
    \label{fig:main_results}
\end{figure}

\section{Discussion} \label{sec:discussion}

In this work, we developed an open-source Jax implementation of a QML framework for medical imaging that enables long-tailed multi-label classification of CXRs. Although we find that the DQC models do not outperform the CDL models on our selected hyperparameters, the performance difference was $<10\%$ across all our experimental setups on both internal and external test sets. We hope to shorten this gap with improvements in theoretical understanding and empirical utility of QML models.

Our benchmarks indicate that our Jax-based implementation is faster than functionally equivalent PyTorch and TensorFlow implementations for DQCs. If we further extrapolate from our benchmarks, a 19-qubit DQC, which takes over 24 hours to train on our hardware with Jax, would take over 48 hours on PyTorch and over 2 weeks on TensorFlow. More generally, the cost of adopting a Jax-based workflow may be easily offset by these runtime improvements.

However, the wall-clock time of DQCs was higher than CDLs due to a slow convergence rate of DQCs with the same learning rate. We observed that the DQC models tended to train up to the maximum epoch limit. This indicates that the DQCs did not end up in a barren plateau in the loss landscape, otherwise they would have stopped earlier. The models that did stop early likely finished in a local minimum instead of a global minimum. It is possible that with more training epochs, the performance difference between CDL and DQC for CXR-14 and CXR-19 could narrow further.

Given the high runtime and slow convergence rate of DQCs, further scalability improvements are critical to follow-up experiments. Parameter initialization could be optimized to enable a more favorable initial training landscape \cite{kashif2023alleviating}. While Adam is a reliable optimizer, other optimizers could potentially provide lower memory usage and faster convergence \cite{anil2019memory,chen2024symbolic,liu2023sophia,ginsburg2019stochastic}. While our batch size was chosen to fit models on a single GPU, the batch size could be tuned to larger sizes through gradient accumulation. Mixed precision training could also lower memory requirements and shorten training time. Finally, using quantum simulation frameworks with additional optimizations could result in further speedups \cite{luo2020yao, wang2022quantumnas, zhang2023tensorcircuit, vallero2024state, pennylane_catalyst}.

For model performance, we observe that both CDL and QDC models were trainable and performed better than random guessing on every label, but the DQC models fail to outperform CDL on every label. This performance deviation could be due to model overparameterization. While the VQC adds a few parameters, the largest contribution of parameters comes from the postprocessing classical layer. ResNet in particular is already overparameterized for medical imaging \cite{holste2023does}, so it is not guaranteed that adding additional parameters would improve performance. It is also possible that quantum circuits could bottleneck the convergence rate or induce a regularizing effect during model training due to limitations in data embedding and measurement layers. This could result in ``quantum utility'' in other areas of machine learning and is an open area of research.

Interestingly, the performance gap in mean AUROC between the CDL and DQC models narrows when testing on MIMIC compared to NIH. We observed that more of the per-label mean AUROC comparisons resulted in non-significant differences between CDL and DQC models. There are two potential confounders in the relatively better performance of DQCs on MIMIC. First, poor generalizability across both models was observed, which limits the applicability of generalization-based arguments. Second, the demographic distributions between NIH and MIMIC are known to differ. Thus, the DQC may have learned certain demographic features unrelated to disease states that are present in higher frequencies in MIMIC vs NIH.

Our volcano plots indicate that labels with a higher occurrence in a data distribution may exhibit more consistent performance trends across different sizes of models and classification tasks. However, these trends may be influenced by demographic or imaging factors specific to each test set, thus it is likely the unique imaging features of each label that contribute to the individual classification challenges of each label. Interestingly, DQC did not outperform CDL in cardiomegaly and pneumonia, two diseases that have been empirically successful with binary classification and QML \cite{decoodt2023hybrid,kulkarni2023classical}. This may be due to differences in the model architectures. It's possible that tuning the learning rate and number of layers, using alternative backbones such as vision transformers \cite{dosovitskiy2020image}, and exploring alternative data encoding methods and quantum circuit ansatz could further improve classification performance for QML models.

Our results point to the need for hyperparameter exploration to better understand the performance range and sensitivity across hyperparameters. While it appears intuitive to use the same hyperparameters for evaluating QML and CML models in the name of fairness, based on our results, CML and QML models are not guaranteed to perform the same due to fundamental differences between quantum and classical computation.
Additional work is needed to determine if the larger variance of DQC performance is an experimental artifact or a fundamental property of DQC models. These follow-up experiments are essential and nontrivial given the large search space of hyperparameters \cite{bowles2024better,dahl2023benchmarking}.

There are certain limitations to our work. 1) We focus on ideal quantum simulation without the presence of noise. 2) We optimize our implementation to fit on a single GPU instead of multiple GPUs. 3) We do not tune model hyperparameters. 4) We focus on the more established Pennylane simulator over newer simulators. We plan to explore these areas in future work.

\section{Conclusion}

Our novel Jax-based quantum machine learning framework for long-tailed chest X-ray classification is scalable for up to 19 qubits and disease labels. Our work is an exciting step towards reducing the computational barrier to quantum machine learning research for medical imaging applications, and opens the door to both further runtime and performance optimizations and characterization of QML behavior across the hyperparameter space.

\bibliographystyle{IEEEtran}
\bibliography{quantum_cxr,quantum_cxr_nogscholar}

\begin{thebibliography}{10}
\providecommand{\url}[1]{#1}
\csname url@samestyle\endcsname
\providecommand{\newblock}{\relax}
\providecommand{\bibinfo}[2]{#2}
\providecommand{\BIBentrySTDinterwordspacing}{\spaceskip=0pt\relax}
\providecommand{\BIBentryALTinterwordstretchfactor}{4}
\providecommand{\BIBentryALTinterwordspacing}{\spaceskip=\fontdimen2\font plus
\BIBentryALTinterwordstretchfactor\fontdimen3\font minus \fontdimen4\font\relax}
\providecommand{\BIBforeignlanguage}[2]{{%
\expandafter\ifx\csname l@#1\endcsname\relax
\typeout{** WARNING: IEEEtran.bst: No hyphenation pattern has been}%
\typeout{** loaded for the language `#1'. Using the pattern for}%
\typeout{** the default language instead.}%
\else
\language=\csname l@#1\endcsname
\fi
#2}}
\providecommand{\BIBdecl}{\relax}
\BIBdecl

\bibitem{nmi2023seeking}
{Nature Machine Intelligence}, ``Seeking a quantum advantage for machine learning,'' \emph{Nat Mach Intell}, vol.~5, no.~8, pp. 813--813, 2023.

\bibitem{caro2022generalization}
M.~C. Caro, H.-Y. Huang, M.~Cerezo, K.~Sharma, A.~Sornborger, L.~Cincio, and P.~J. Coles, ``Generalization in quantum machine learning from few training data,'' \emph{Nature communications}, vol.~13, no.~1, p. 4919, 2022.

\bibitem{abbas2021power}
A.~Abbas, D.~Sutter, C.~Zoufal, A.~Lucchi, A.~Figalli, and S.~Woerner, ``The power of quantum neural networks,'' \emph{Nature Computational Science}, vol.~1, no.~6, pp. 403--409, 2021.

\bibitem{huang2021power}
H.-Y. Huang, M.~Broughton, M.~Mohseni, R.~Babbush, S.~Boixo, H.~Neven, and J.~R. McClean, ``Power of data in quantum machine learning,'' \emph{Nature communications}, vol.~12, no.~1, p. 2631, 2021.

\bibitem{jia2023importance}
Z.~Jia, J.~Chen, X.~Xu, J.~Kheir, J.~Hu, H.~Xiao, S.~Peng, X.~S. Hu, D.~Chen, and Y.~Shi, ``The importance of resource awareness in artificial intelligence for healthcare,'' \emph{Nature Machine Intelligence}, vol.~5, no.~7, pp. 687--698, 2023.

\bibitem{holste2022long}
G.~Holste, S.~Wang, Z.~Jiang, T.~C. Shen, G.~Shih, R.~M. Summers, Y.~Peng, and Z.~Wang, ``Long-tailed classification of thorax diseases on chest x-ray: A new benchmark study,'' in \emph{MICCAI Workshop on Data Augmentation, Labelling, and Imperfections}.\hskip 1em plus 0.5em minus 0.4em\relax Springer, 2022, pp. 22--32.

\bibitem{decoodt2023hybrid}
P.~Decoodt, T.~J. Liang, S.~Bopardikar, H.~Santhanam, A.~Eyembe, B.~Garcia-Zapirain, and D.~Sierra-Sosa, ``Hybrid classical--quantum transfer learning for cardiomegaly detection in chest x-rays,'' \emph{Journal of Imaging}, vol.~9, no.~7, p. 128, 2023.

\bibitem{kulkarni2023classical}
V.~Kulkarni, S.~Pawale, and A.~Kharat, ``A classical--quantum convolutional neural network for detecting pneumonia from chest radiographs,'' \emph{Neural Computing and Applications}, vol.~35, no.~21, pp. 15\,503--15\,510, 2023.

\bibitem{houssein2022hybrid}
E.~H. Houssein, Z.~Abohashima, M.~Elhoseny, and W.~M. Mohamed, ``Hybrid quantum-classical convolutional neural network model for covid-19 prediction using chest x-ray images,'' \emph{Journal of Computational Design and Engineering}, vol.~9, no.~2, pp. 343--363, 2022.

\bibitem{bowles2024better}
J.~Bowles, S.~Ahmed, and M.~Schuld, ``Better than classical? the subtle art of benchmarking quantum machine learning models,'' \emph{arXiv preprint arXiv:2403.07059}, 2024.

\bibitem{asadi2024hybrid}
A.~Asadi, A.~Dusko, C.-Y. Park, V.~Michaud-Rioux, I.~Schoch, S.~Shu, T.~Vincent, and L.~J. O'Riordan, ``Hybrid quantum programming with pennylane lightning on hpc platforms,'' \emph{arXiv preprint arXiv:2403.02512}, 2024.

\bibitem{schuld2022quantum}
M.~Schuld and N.~Killoran, ``Is quantum advantage the right goal for quantum machine learning?'' \emph{Prx Quantum}, vol.~3, no.~3, p. 030101, 2022.

\bibitem{arunachalam2018optimal}
S.~Arunachalam and R.~De~Wolf, ``Optimal quantum sample complexity of learning algorithms,'' \emph{Journal of Machine Learning Research}, vol.~19, no.~71, pp. 1--36, 2018.

\bibitem{tang2022dequantizing}
E.~Tang, ``Dequantizing algorithms to understand quantum advantage in machine learning,'' \emph{Nature Reviews Physics}, vol.~4, no.~11, pp. 692--693, 2022.

\bibitem{gil2024understanding}
E.~Gil-Fuster, J.~Eisert, and C.~Bravo-Prieto, ``Understanding quantum machine learning also requires rethinking generalization,'' \emph{Nature Communications}, vol.~15, no.~1, pp. 1--12, 2024.

\bibitem{holmes2022connecting}
Z.~Holmes, K.~Sharma, M.~Cerezo, and P.~J. Coles, ``Connecting ansatz expressibility to gradient magnitudes and barren plateaus,'' \emph{PRX Quantum}, vol.~3, no.~1, p. 010313, 2022.

\bibitem{ccalli2021deep}
E.~{\c{C}}all{\i}, E.~Sogancioglu, B.~van Ginneken, K.~G. van Leeuwen, and K.~Murphy, ``Deep learning for chest x-ray analysis: A survey,'' \emph{Medical Image Analysis}, vol.~72, p. 102125, 2021.

\bibitem{wang2017chestx}
X.~Wang, Y.~Peng, L.~Lu, Z.~Lu, M.~Bagheri, and R.~M. Summers, ``Chestx-ray8: Hospital-scale chest x-ray database and benchmarks on weakly-supervised classification and localization of common thorax diseases,'' in \emph{Proceedings of the IEEE conference on computer vision and pattern recognition}, 2017, pp. 2097--2106.

\bibitem{irvin2019chexpert}
J.~Irvin, P.~Rajpurkar, M.~Ko, Y.~Yu, S.~Ciurea-Ilcus, C.~Chute, H.~Marklund, B.~Haghgoo, R.~Ball, K.~Shpanskaya \emph{et~al.}, ``Chexpert: A large chest radiograph dataset with uncertainty labels and expert comparison,'' in \emph{Proceedings of the AAAI conference on artificial intelligence}, vol.~33, no.~01, 2019, pp. 590--597.

\bibitem{johnson2019mimic}
A.~E. Johnson, T.~J. Pollard, N.~R. Greenbaum, M.~P. Lungren, C.-y. Deng, Y.~Peng, Z.~Lu, R.~G. Mark, S.~J. Berkowitz, and S.~Horng, ``Mimic-cxr-jpg, a large publicly available database of labeled chest radiographs,'' \emph{arXiv preprint arXiv:1901.07042}, 2019.

\bibitem{bustos2020padchest}
A.~Bustos, A.~Pertusa, J.-M. Salinas, and M.~De~La Iglesia-Vaya, ``Padchest: A large chest x-ray image dataset with multi-label annotated reports,'' \emph{Medical image analysis}, vol.~66, p. 101797, 2020.

\bibitem{sunkel2023hybrid}
L.~S{\"u}nkel, D.~Martyniuk, J.~J. Reichwald, A.~Morariu, R.~H. Seggoju, P.~Altmann, C.~Roch, and A.~Paschke, ``Hybrid quantum machine learning assisted classification of covid-19 from computed tomography scans,'' in \emph{2023 IEEE International Conference on Quantum Computing and Engineering (QCE)}, vol.~1.\hskip 1em plus 0.5em minus 0.4em\relax IEEE, 2023, pp. 356--366.

\bibitem{duan2020survey}
B.~Duan, J.~Yuan, C.-H. Yu, J.~Huang, and C.-Y. Hsieh, ``A survey on hhl algorithm: From theory to application in quantum machine learning,'' \emph{Physics Letters A}, vol. 384, no.~24, p. 126595, 2020.

\bibitem{liu2024towards}
J.~Liu, M.~Liu, J.-P. Liu, Z.~Ye, Y.~Wang, Y.~Alexeev, J.~Eisert, and L.~Jiang, ``Towards provably efficient quantum algorithms for large-scale machine-learning models,'' \emph{Nature Communications}, vol.~15, no.~1, p. 434, 2024.

\bibitem{havlivcek2019supervised}
V.~Havl{\'\i}{\v{c}}ek, A.~D. C{\'o}rcoles, K.~Temme, A.~W. Harrow, A.~Kandala, J.~M. Chow, and J.~M. Gambetta, ``Supervised learning with quantum-enhanced feature spaces,'' \emph{Nature}, vol. 567, no. 7747, pp. 209--212, 2019.

\bibitem{dallaire2018quantum}
P.-L. Dallaire-Demers and N.~Killoran, ``Quantum generative adversarial networks,'' \emph{Physical Review A}, vol.~98, no.~1, p. 012324, 2018.

\bibitem{amin2018quantum}
M.~H. Amin, E.~Andriyash, J.~Rolfe, B.~Kulchytskyy, and R.~Melko, ``Quantum boltzmann machine,'' \emph{Physical Review X}, vol.~8, no.~2, p. 021050, 2018.

\bibitem{cerezo2021variational}
M.~Cerezo, A.~Arrasmith, R.~Babbush, S.~C. Benjamin, S.~Endo, K.~Fujii, J.~R. McClean, K.~Mitarai, X.~Yuan, L.~Cincio \emph{et~al.}, ``Variational quantum algorithms,'' \emph{Nature Reviews Physics}, vol.~3, no.~9, pp. 625--644, 2021.

\bibitem{farhi2018classification}
E.~Farhi and H.~Neven, ``Classification with quantum neural networks on near term processors,'' \emph{arXiv preprint arXiv:1802.06002}, 2018.

\bibitem{mcclean2018barren}
J.~R. McClean, S.~Boixo, V.~N. Smelyanskiy, R.~Babbush, and H.~Neven, ``Barren plateaus in quantum neural network training landscapes,'' \emph{Nature communications}, vol.~9, no.~1, p. 4812, 2018.

\bibitem{sim2019expressibility}
S.~Sim, P.~D. Johnson, and A.~Aspuru-Guzik, ``Expressibility and entangling capability of parameterized quantum circuits for hybrid quantum-classical algorithms,'' \emph{Advanced Quantum Technologies}, vol.~2, no.~12, p. 1900070, 2019.

\bibitem{friedrich2022avoiding}
L.~Friedrich and J.~Maziero, ``Avoiding barren plateaus with classical deep neural networks,'' \emph{Physical Review A}, vol. 106, no.~4, p. 042433, 2022.

\bibitem{mari2020transfer}
A.~Mari, T.~R. Bromley, J.~Izaac, M.~Schuld, and N.~Killoran, ``Transfer learning in hybrid classical-quantum neural networks,'' \emph{Quantum}, vol.~4, p. 340, 2020.

\bibitem{chen2021end}
S.~Y.-C. Chen, C.-M. Huang, C.-W. Hsing, and Y.-J. Kao, ``An end-to-end trainable hybrid classical-quantum classifier,'' \emph{Machine Learning: Science and Technology}, vol.~2, no.~4, p. 045021, 2021.

\bibitem{stilck2021limitations}
D.~Stilck~Fran{\c{c}}a and R.~Garcia-Patron, ``Limitations of optimization algorithms on noisy quantum devices,'' \emph{Nature Physics}, vol.~17, no.~11, pp. 1221--1227, 2021.

\bibitem{wang2021noise}
S.~Wang, E.~Fontana, M.~Cerezo, K.~Sharma, A.~Sone, L.~Cincio, and P.~J. Coles, ``Noise-induced barren plateaus in variational quantum algorithms,'' \emph{Nature communications}, vol.~12, no.~1, p. 6961, 2021.

\bibitem{gottesman2022opportunities}
D.~Gottesman, ``Opportunities and challenges in fault-tolerant quantum computation,'' \emph{arXiv preprint arXiv:2210.15844}, 2022.

\bibitem{quek2022exponentially}
Y.~Quek, D.~S. Fran{\c{c}}a, S.~Khatri, J.~J. Meyer, and J.~Eisert, ``Exponentially tighter bounds on limitations of quantum error mitigation,'' \emph{arXiv preprint arXiv:2210.11505}, 2022.

\bibitem{zhou2020limits}
Y.~Zhou, E.~M. Stoudenmire, and X.~Waintal, ``What limits the simulation of quantum computers?'' \emph{Physical Review X}, vol.~10, no.~4, p. 041038, 2020.

\bibitem{endo2021hybrid}
S.~Endo, Z.~Cai, S.~C. Benjamin, and X.~Yuan, ``Hybrid quantum-classical algorithms and quantum error mitigation,'' \emph{Journal of the Physical Society of Japan}, vol.~90, no.~3, p. 032001, 2021.

\bibitem{vallero2024state}
M.~Vallero, F.~Vella, and P.~Rech, ``State of practice: evaluating gpu performance of state vector and tensor network methods,'' \emph{arXiv preprint arXiv:2401.06188}, 2024.

\bibitem{kyriienko2022unsupervised}
O.~Kyriienko and E.~B. Magnusson, ``Unsupervised quantum machine learning for fraud detection,'' \emph{arXiv preprint arXiv:2208.01203}, 2022.

\bibitem{gianelle2022quantum}
A.~Gianelle, P.~Koppenburg, D.~Lucchesi, D.~Nicotra, E.~Rodrigues, L.~Sestini, J.~de~Vries, and D.~Zuliani, ``Quantum machine learning for b-jet charge identification,'' \emph{Journal of High Energy Physics}, vol. 2022, no.~8, pp. 1--24, 2022.

\bibitem{wolf2019huggingface}
T.~Wolf, L.~Debut, V.~Sanh, J.~Chaumond, C.~Delangue, A.~Moi, P.~Cistac, T.~Rault, R.~Louf, M.~Funtowicz \emph{et~al.}, ``Huggingface's transformers: State-of-the-art natural language processing,'' \emph{arXiv preprint arXiv:1910.03771}, 2019.

\bibitem{azevedo2022quantum}
V.~Azevedo, C.~Silva, and I.~Dutra, ``Quantum transfer learning for breast cancer detection,'' \emph{Quantum Machine Intelligence}, vol.~4, no.~1, p.~5, 2022.

\bibitem{bradbury2018jax}
J.~Bradbury, R.~Frostig, P.~Hawkins, M.~J. Johnson, C.~Leary, D.~Maclaurin, G.~Necula, A.~Paszke, J.~VanderPlas, S.~Wanderman-Milne \emph{et~al.}, ``Jax: composable transformations of python+ numpy programs,'' 2018.

\bibitem{bergholm2018pennylane}
V.~Bergholm, J.~Izaac, M.~Schuld, C.~Gogolin, S.~Ahmed, V.~Ajith, M.~S. Alam, G.~Alonso-Linaje, B.~AkashNarayanan, A.~Asadi \emph{et~al.}, ``Pennylane: Automatic differentiation of hybrid quantum-classical computations,'' \emph{arXiv preprint arXiv:1811.04968}, 2018.

\bibitem{heek2020flax}
J.~Heek, A.~Levskaya, A.~Oliver, M.~Ritter, B.~Rondepierre, A.~Steiner, and M.~van Zee, ``Flax: A neural network library and ecosystem for jax,'' \emph{Version 0.3}, vol.~3, pp. 14--26, 2020.

\bibitem{deepmind2020jax}
\BIBentryALTinterwordspacing
DeepMind, I.~Babuschkin, K.~Baumli, A.~Bell, S.~Bhupatiraju, J.~Bruce, P.~Buchlovsky, D.~Budden, T.~Cai, A.~Clark, I.~Danihelka, A.~Dedieu, C.~Fantacci, J.~Godwin, C.~Jones, R.~Hemsley, T.~Hennigan, M.~Hessel, S.~Hou, S.~Kapturowski, T.~Keck, I.~Kemaev, M.~King, M.~Kunesch, L.~Martens, H.~Merzic, V.~Mikulik, T.~Norman, G.~Papamakarios, J.~Quan, R.~Ring, F.~Ruiz, A.~Sanchez, L.~Sartran, R.~Schneider, E.~Sezener, S.~Spencer, S.~Srinivasan, M.~Stanojevi\'{c}, W.~Stokowiec, L.~Wang, G.~Zhou, and F.~Viola, ``The {D}eep{M}ind {JAX} {E}cosystem,'' 2020. [Online]. Available: \url{http://github.com/google-deepmind}
\BIBentrySTDinterwordspacing

\bibitem{patry_safetensors}
\BIBentryALTinterwordspacing
N.~Patry, ``{Safetensors}: Simple, safe way to store and distribute tensors.'' [Online]. Available: \url{https://github.com/huggingface/safetensors}
\BIBentrySTDinterwordspacing

\bibitem{collet2018zstandard}
Y.~Collet and M.~Kucherawy, ``Zstandard compression and the application/zstd media type,'' 2018.

\bibitem{jenks_diskcache}
\BIBentryALTinterwordspacing
G.~Jenks, ``{DiskCache: Disk Backed Cache}.'' [Online]. Available: \url{https://github.com/grantjenks/python-diskcache}
\BIBentrySTDinterwordspacing

\bibitem{joblib_joblib}
\BIBentryALTinterwordspacing
J.~development team, ``{Joblib: running Python functions as pipeline jobs}.'' [Online]. Available: \url{https://github.com/joblib/joblib}
\BIBentrySTDinterwordspacing

\bibitem{klambauer2017self}
G.~Klambauer, T.~Unterthiner, A.~Mayr, and S.~Hochreiter, ``Self-normalizing neural networks,'' \emph{Advances in neural information processing systems}, vol.~30, 2017.

\bibitem{holste2023does}
G.~Holste, Z.~Jiang, A.~Jaiswal, M.~Hanna, S.~Minkowitz, A.~C. Legasto, J.~G. Escalon, S.~Steinberger, M.~Bittman, T.~C. Shen \emph{et~al.}, ``How does pruning impact long-tailed multi-label medical image classifiers?'' in \emph{International Conference on Medical Image Computing and Computer-Assisted Intervention}.\hskip 1em plus 0.5em minus 0.4em\relax Springer, 2023, pp. 663--673.

\bibitem{kashif2023alleviating}
M.~Kashif, M.~Rashid, S.~Al-Kuwari, and M.~Shafique, ``Alleviating barren plateaus in parameterized quantum machine learning circuits: Investigating advanced parameter initialization strategies,'' \emph{arXiv preprint arXiv:2311.13218}, 2023.

\bibitem{anil2019memory}
R.~Anil, V.~Gupta, T.~Koren, and Y.~Singer, ``Memory efficient adaptive optimization,'' \emph{Advances in Neural Information Processing Systems}, vol.~32, 2019.

\bibitem{chen2024symbolic}
X.~Chen, C.~Liang, D.~Huang, E.~Real, K.~Wang, H.~Pham, X.~Dong, T.~Luong, C.-J. Hsieh, Y.~Lu \emph{et~al.}, ``Symbolic discovery of optimization algorithms,'' \emph{Advances in Neural Information Processing Systems}, vol.~36, 2024.

\bibitem{liu2023sophia}
H.~Liu, Z.~Li, D.~Hall, P.~Liang, and T.~Ma, ``Sophia: A scalable stochastic second-order optimizer for language model pre-training,'' \emph{arXiv preprint arXiv:2305.14342}, 2023.

\bibitem{ginsburg2019stochastic}
B.~Ginsburg, P.~Castonguay, O.~Hrinchuk, O.~Kuchaiev, V.~Lavrukhin, R.~Leary, J.~Li, H.~Nguyen, Y.~Zhang, and J.~M. Cohen, ``Stochastic gradient methods with layer-wise adaptive moments for training of deep networks,'' \emph{arXiv preprint arXiv:1905.11286}, 2019.

\bibitem{luo2020yao}
X.-Z. Luo, J.-G. Liu, P.~Zhang, and L.~Wang, ``Yao. jl: Extensible, efficient framework for quantum algorithm design,'' \emph{Quantum}, vol.~4, p. 341, 2020.

\bibitem{wang2022quantumnas}
H.~Wang, Y.~Ding, J.~Gu, Y.~Lin, D.~Z. Pan, F.~T. Chong, and S.~Han, ``Quantumnas: Noise-adaptive search for robust quantum circuits,'' in \emph{2022 IEEE International Symposium on High-Performance Computer Architecture (HPCA)}.\hskip 1em plus 0.5em minus 0.4em\relax IEEE, 2022, pp. 692--708.

\bibitem{zhang2023tensorcircuit}
S.-X. Zhang, J.~Allcock, Z.-Q. Wan, S.~Liu, J.~Sun, H.~Yu, X.-H. Yang, J.~Qiu, Z.~Ye, Y.-Q. Chen \emph{et~al.}, ``Tensorcircuit: a quantum software framework for the nisq era,'' \emph{Quantum}, vol.~7, p. 912, 2023.

\bibitem{pennylane_catalyst}
\BIBentryALTinterwordspacing
P.~development team, ``{Catalyst: A JIT compiler for hybrid quantum programs in PennyLane }.'' [Online]. Available: \url{https://github.com/PennyLaneAI/catalyst}
\BIBentrySTDinterwordspacing

\bibitem{dosovitskiy2020image}
A.~Dosovitskiy, L.~Beyer, A.~Kolesnikov, D.~Weissenborn, X.~Zhai, T.~Unterthiner, M.~Dehghani, M.~Minderer, G.~Heigold, S.~Gelly \emph{et~al.}, ``An image is worth 16x16 words: Transformers for image recognition at scale,'' \emph{arXiv preprint arXiv:2010.11929}, 2020.

\bibitem{dahl2023benchmarking}
G.~E. Dahl, F.~Schneider, Z.~Nado, N.~Agarwal, C.~S. Sastry, P.~Hennig, S.~Medapati, R.~Eschenhagen, P.~Kasimbeg, D.~Suo \emph{et~al.}, ``Benchmarking neural network training algorithms,'' \emph{arXiv preprint arXiv:2306.07179}, 2023.

\end{thebibliography}

\end{document}